\documentclass{acmart}
\usepackage{makecell}
\usepackage{url}
\usepackage{graphicx}
\usepackage{booktabs}
\usepackage{CJKutf8}
\usepackage{subcaption}
\usepackage{float}
\usepackage{amsmath}
\usepackage{multirow}
\usepackage{booktabs}
\usepackage{url}
\usepackage{array}
\usepackage{enumitem}
\usepackage{pifont}
\usepackage{bm}
\usepackage{lipsum}
\usepackage{CJKutf8}
\usepackage{tikz}
\usepackage[edges]{forest}
\usepackage{wrapfig}
\usepackage{subcaption}
\usepackage{natbib}
\definecolor{hidden-draw}{RGB}{20,68,106}
\definecolor{hidden-pink}{RGB}{255,245,247}
\usepackage[capitalize]{cleveref}
\usepackage{multirow}
\usepackage[normalem]{ulem}

\AtBeginDocument{%
  }

\setcopyright{acmlicensed}
\copyrightyear{2026}
\acmYear{2026}
\acmDOI{XXXXXXX.XXXXXXX}

\acmISBN{978-1-4503-XXXX-X/18/06}

\begin{document}

\title{Alignment of Diffusion Models: Fundamentals, Challenges, and Future}

\author{Buhua Liu}
\affiliation{%
  \institution{The Hong Kong University of Science and Technology (Guangzhou)}
  \city{Guangzhou}
  \state{Guangdong}
  \country{China}
}
\email{bryceliu@foxmail.com}

\author{Shitong Shao}
\affiliation{%
  \institution{The Hong Kong University of Science and Technology (Guangzhou)}
  \city{Guangzhou}
  \state{Guangdong}
  \country{China}
}
\email{sshao213@connect.hkust-gz.edu.cn}

\author{Bao Li}
\affiliation{%
  \institution{Institute of Automation, Chinese Academy of Sciences}
  \city{Beijing}
  \country{China}
}
\email{libao2023@gmail.com}

\author{Lichen Bai}
\affiliation{%
  \institution{The Hong Kong University of Science and Technology (Guangzhou)}
  \city{Guangzhou}
  \state{Guangdong}
  \country{China}
}
\email{lichenbai@hkust-gz.edu.cn}

\author{Zhiqiang Xu}
\affiliation{%
  \institution{Mohamed bin Zayed University of Artificial Intelligence}
  \city{Masdar City}
  \state{Abu Dhabi}
  \country{UAE}
}
\email{Zhiqiang.Xu@mbzuai.ac.ae}

\author{Haoyi Xiong}
\affiliation{%
  \institution{Baidu Inc.}
  \city{Beijing}
  \country{China}
}
\email{xhyccc@gmail.com}

\author{James Kwok}
\affiliation{%
  \institution{The Hong Kong University of Science and Technology}
  \country{Hong Kong}
}
\email{jamesk@cse.ust.hk}

\author{Sumi Helal}
\affiliation{%
  \institution{The University of Bologna}
  \city{Bologna}
  \state{Emilia-Romagna}
  \country{Italy}
}
\email{sumi.helal@gmail.com}

\author{Zeke Xie}
\authornote{Corresponding author}
\affiliation{%
  \institution{The Hong Kong University of Science and Technology (Guangzhou)}
  \city{Guangzhou}
  \state{Guangdong}
  \country{China}
}
\email{zekexie@hkust-gz.edu.cn}

\renewcommand{\shortauthors}{Buhua Liu, Shitong Shao, Bao Li, Lichen Bai, Zhiqiang Xu, Haoyi Xiong, James Kwok, Sumi Helal, Zeke Xie}

\begin{abstract}
  Diffusion models have emerged as the leading paradigm in generative modeling, excelling in various applications. Despite their success, these models often misalign with human intentions and generate results with undesired properties or even harmful content. Inspired by the success and popularity of alignment in tuning large language models, recent studies have investigated aligning diffusion models with human expectations and preferences. This work mainly reviews alignment of diffusion models, covering advancements in fundamentals of alignment, alignment techniques of diffusion models, preference benchmarks, and evaluation for diffusion models. Moreover, we discuss key perspectives on current challenges and promising future directions on solving the remaining challenges in alignment of diffusion models. To the best of our knowledge, our work is the first comprehensive review paper for researchers and engineers to comprehend, practice, and research alignment of diffusion models.
\end{abstract}




\maketitle
\section{Introduction}
\label{sec:intro} 
\begin{figure}[htbp]
    \centering
    \begin{subfigure}[b]{0.58\textwidth}
        \centering
        \includegraphics[width=\textwidth]{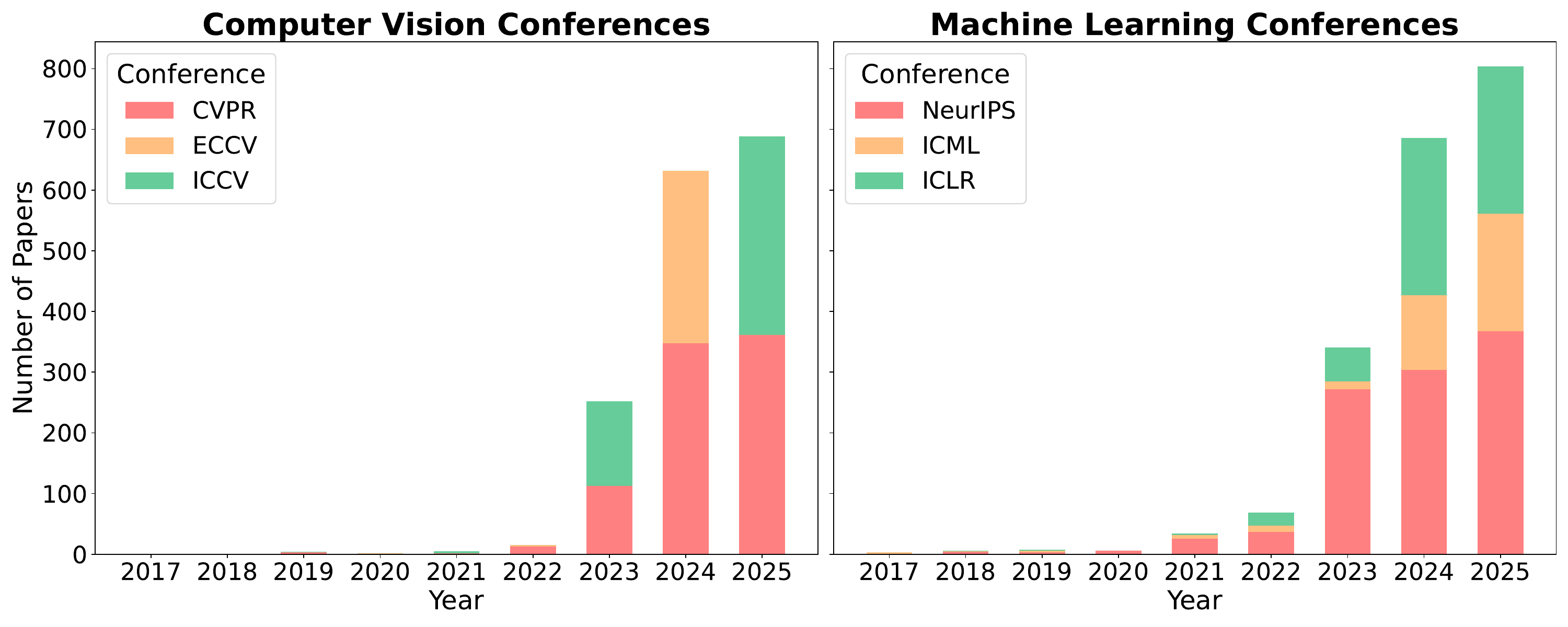}
        \caption{Diffusion model paper trends at top conferences.}
        \label{fig:intro_stats_conference}
        \Description{The trend in the number of papers on diffusion models published in top computer vision conferences (CVPR, ECCV, ICCV) (left) and top machine learning conferences (NeurIPS, ICML, ICLR) (right) in recent years, which shows the growing interest in diffusion models at top conferences.}
    \end{subfigure}
    \begin{subfigure}[b]{0.4\textwidth}
        \centering
        \includegraphics[width=\textwidth]{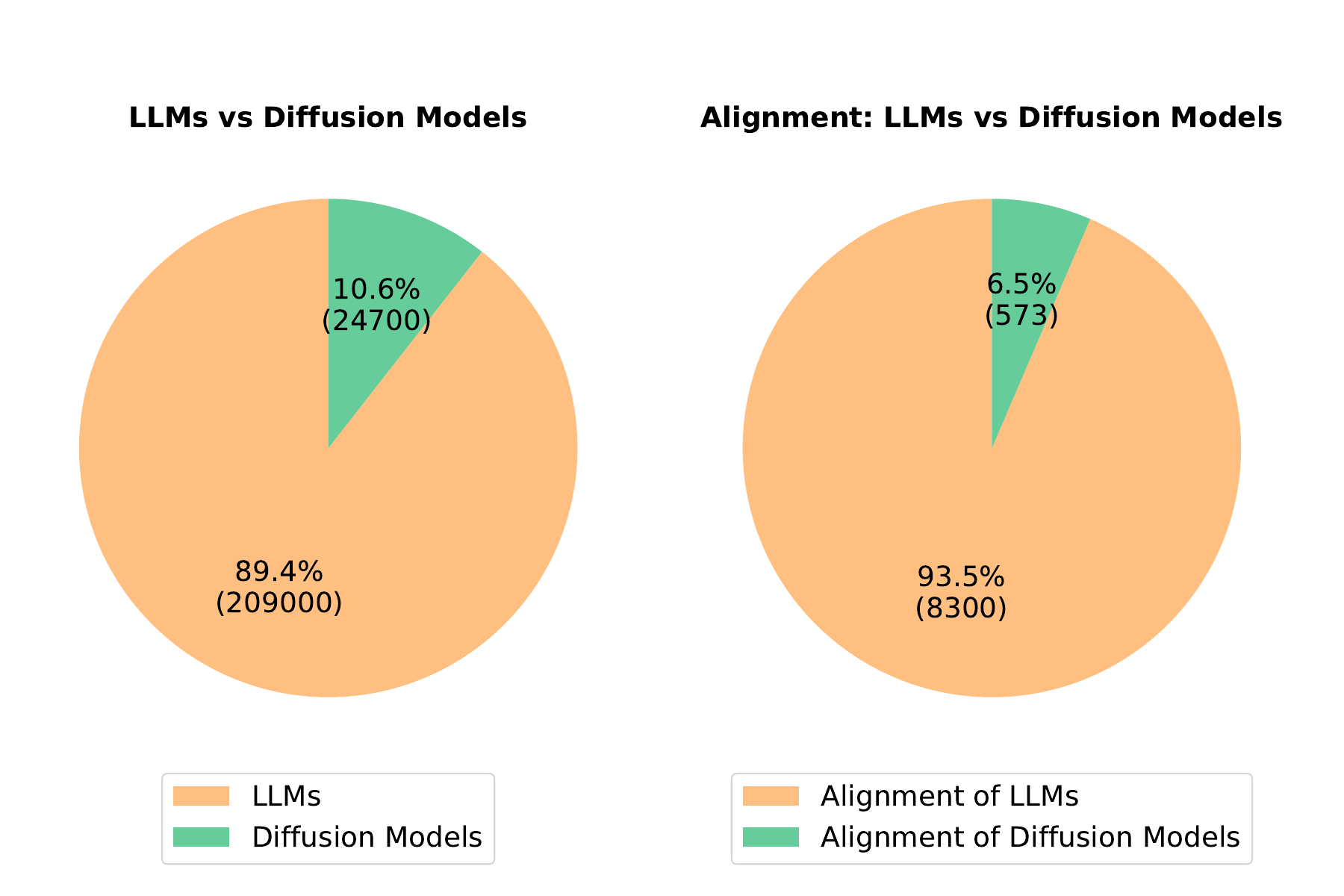}
        \caption{Research focus: LLMs vs. Diffusion Models and their alignment.}
        \label{fig:intro_stats_pie}
        \Description{The left pie chart illustrates the distribution of total papers, showing that LLMs account for 89.4\% of the research (209,000 papers), compared to 10.6\% (24,700 papers) for diffusion models. The right pie chart highlights alignment research, revealing a significant skew towards LLMs, with 93.5\% of alignment-related studies (8,300 papers) focused on LLMs, while only 6.5\% (573 papers) address alignment in diffusion models. This disparity underscores the relatively nascent focus on alignment within diffusion models compared to the research on LLMs.}
    \end{subfigure}
    \caption{Statistical overview of research trends. (a) The number of papers on diffusion models at top computer vision conferences (CVPR, ECCV, ICCV) and top machine learning conferences (NeurIPS, ICML, ICLR) since 2017, indicating a growing interest. Note that ECCV and ICCV are held biennially. (b) Comparison of the ratio of papers on LLMs vs. diffusion models (left pie) and the research focus on alignment within these areas (right pie), highlighting the nascent stage of diffusion model alignment.}
    \label{fig:intro_stats}
\end{figure}

Diffusion models~\citep{sohl2015deep,ho2020denoising,song2021scorebased}
have emerged as the dominant paradigm, surpassing previous state-of-the-art generative models such as generative adversarial networks (GANs)~\citep{goodfellow2020generative,dhariwal2021diffusion,10.1145/3459992,10.1145/3446374,10.1145/3439723,10.1145/3463475} and variational autoencoders (VAEs)~\citep{kingma2013auto}. Diffusion models have demonstrated the impressive performance and success in various generative tasks, including image generation~\citep{esser2024scaling,10.1145/3665869}, video generation~\citep{ho2022imagen,blattmann2023align}, text generation~\citep{lou2024discrete}, audio synthesis~\citep{kong2020diffwave,huang2023make}, 3D generation~\citep{ye2024dreamreward,10.1145/3626193}, music generation~\citep{10.1145/3672554}, and molecule generation~\citep{xu2022geodiff,gu2024aligning}. {\cref{fig:intro_stats_conference} illustrates the trend in the number of papers on diffusion models published in top computer vision conferences (CVPR, ICCV, ECCV) and top machine learning conferences (NeurIPS, ICML, ICLR) in recent years, highlighting the growing interest in diffusion models at these leading conferences.}

\begin{figure}[htbp]
    \centering
    \includegraphics[width=0.8\textwidth]{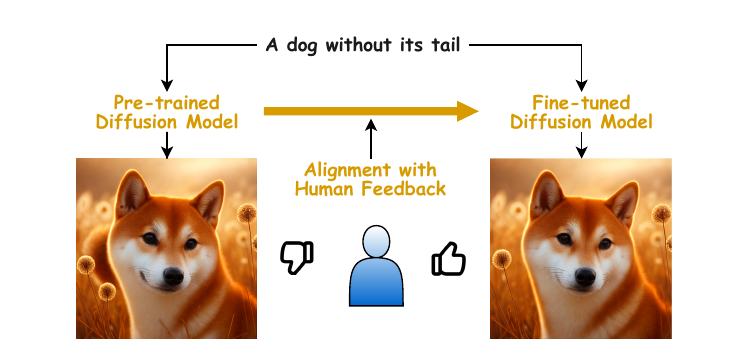}
    \caption{Conceptual illustration of diffusion model misalignment and the goal of alignment. A pre-trained model may generate an output that deviates from human intentions or desired qualities (e.g., missing details in a prompt). An aligned model, after an alignment process, produces an output that better reflects human preferences for the same input.}
    \label{fig:alignment_illustration}
    \Description{A two-panel diagram illustrating model alignment. Panel on the left, titled 'Pre-trained Model Output (Misaligned)', depicts a generated image that does not accurately match an example prompt (e.g., 'a cat wearing a hat'). Panel on the right, titled 'Aligned Model Output', shows an image that accurately depicts 'a cat wearing a hat', demonstrating successful alignment.}
\end{figure}

However, the diffusion training objective does not necessarily align well with human intentions and preferences. For instance, images generated by pre-trained text-to-image (T2I) models may {generate images that, while technically plausible, may fail to capture specific artistic nuances or accurately represent complex textual descriptions}~\citep{lee2023aligning,blacktraining,fan2023reinforcement}{, which can hinder practical applications or diminish user satisfaction}. Similarly, in drug discovery, pre-trained diffusion models typically lack the ability to generate molecules with high binding affinity and structural rationality~\citep{gu2024aligning}, a critical issue that can directly impact therapeutic efficacy. {\cref{fig:alignment_illustration} provides a conceptual illustration of such misalignment.} To address this mismatch, recent works have begun to optimize pre-trained diffusion models directly for human preference or certain desired properties, aiming to more controllable data generation~\citep{10.1145/3648609} beyond simply modeling the training data distribution. 

Within the community of language modeling, recent powerful large language models (LLMs) like GPT-4~\citep{achiam2023gpt}, Llama~\citep{touvron2023llama,dubey2024llama}, and Qwen 3~\citep{yang2025qwen3} are typically trained in two stages. In the first pre-training stage, they are trained on a vast textual corpus with the objective of predicting the next tokens. In the second post-training stage, they are fine-tuned to follow instructions, align with human preferences, and improve capabilities like coding and factuality. The post-training process usually involves supervised fine-tuning (SFT) followed by alignment with human feedback, using techniques such as reinforcement learning from human feedback (RLHF)~\citep{ouyang2022training,achiam2023gpt}, and direct preference optimization (DPO)~\citep{rafailov2023direct,dubey2024llama}. {The success of these techniques in LLMs is particularly relevant as they demonstrate the feasibility and effectiveness of aligning complex generative models with nuanced human preferences, thereby providing a validated conceptual and methodological roadmap for diffusion models.} LLMs trained using this two-stage process have achieved state-of-the-art performance~\citep{achiam2023gpt,comanici2025gemini} in various language generation tasks and have been deployed in commercial applications such as ChatGPT.

Inspired by the success of aligning LLMs \citep{wang2023aligning}, there is growing interest in better aligning diffusion models with human intentions to enhance their capabilities. \cref{fig:intro_stats_pie} visualizes the paper counts on LLMs and diffusion models, as well as their alignment studies \footnote{Data obtained from Google Scholar as of January 15, 2026. 
}. The left pie chart shows LLMs account for 89.4\% of the studies, while diffusion models account for 10.6\%. The right chart highlights that, at this point, 93.5\% of alignment studies focus on LLMs, while only 6.5\% address diffusion models. This disparity, clearly illustrated in \cref{fig:intro_stats_pie}, not only underscores the relatively nascent stage of alignment research for diffusion models compared to LLMs but also signals a significant opportunity and pressing need for focused investigation in this domain. This survey aims to facilitate such research by consolidating current knowledge and outlining future directions. The fundamental differences between LLMs and diffusion models, where LLMs predict sequences of tokens and diffusion models progressively reverse a noise-adding diffusion process, {and where the continuous, high-dimensional nature of image generation versus discrete token prediction in LLMs presents unique challenges for defining preferences and applying feedback,} along with their uniquely advantageous application domains, such as LLMs for language generation and diffusion models for image generation, make the study of diffusion model alignment an independent and valuable area of interest.
\tikzstyle{my-box}=[   rectangle,
    draw=hidden-draw,
    rounded corners,
    text opacity=1,
    minimum height=1.5em,
    minimum width=5em,
    inner sep=2pt,
    align=center,
    fill opacity=.5,
    line width=0.8pt,
]
\tikzstyle{leaf}=[my-box, minimum height=1.5em,
    fill=hidden-pink!80, text=black, align=left,font=\normalsize,
    inner xsep=2pt,
    inner ysep=4pt,
    line width=0.8pt,
]
\begin{figure*}[!t]
\centering
\resizebox{0.99\linewidth}{!}{
\begin{forest}
    forked edges,
    for tree={
        grow=east,
        reversed=true,
        anchor=base west,
        parent anchor=east,
        child anchor=west,
        base=center,
        font=\footnotesize,
        rectangle,
        draw=hidden-draw,
        rounded corners,
        align=left,
        text centered,
        minimum width=4em,
        edge+={darkgray, line width=1pt},
        s sep=3pt,
        inner xsep=2pt,
        inner ysep=3pt,
        line width=0.8pt,
        ver/.style={rotate=90, child anchor=north, parent anchor=south, anchor=center},
    },
    where level=1{text width=10em,font=\normalsize,}{},
    where level=2{text width=10em,font=\normalsize,}{},
    where level=3{text width=11em,font=\normalsize,}{},
    where level=4{text width=7em,font=\normalsize,}{},
    where level=5{text width=7em,font=\normalsize,}{},
    [
        Alignment of Diffusion Models, ver
        [
            Fundamentals of \\ Human Alignment
            [
                Preference Data \\ and Modeling
                [
                    Bradley-Terry Model ~\citep{bradley1952rank}{,} \\ Plackett-Luce Model ~\citep{plackett1975analysis, luce1959individual}, leaf, text width=18em
                ]
            ]
            [
                Alignment \\ Algorithms
                [
                    Reinforcement Learning \\ from Human Feedback
                    [
                        InstructGPT ~\citep{ouyang2022training}{,}  
                        REINFORCE ~\citep{ahmadian2024back}{,} \\
                        RAFT ~\citep{dong2023raft}{,}
                        RRHF ~\citep{yuan2023rrhf}{,} RSO ~\citep{liu2024statistical}{,} \\
                        CAI ~\citep{bai2022constitutional}{,}
                        RLAIF ~\citep{lee2024rlaif}, leaf, text width=15em
                    ]
                ]
                [
                    Direct Preference \\ Optimization
                    [
                        DPO ~\citep{rafailov2023direct}{,} 
                        IPO ~\citep{azar2024general}{,}
                        ORPO ~\citep{hong2024orpo}{,} \\
                        KTO ~\citep{ethayarajh2024kto}{,}
                        PRO ~\citep{song2024preference}{,} LiPO ~\citep{liu2024lipo}, leaf, text width=15em
                    ]
                ]
            ]
        ]
        [
            Human Alignment of \\ Diffusion Models
            [
                Training-based \\ Alignment
                [
                    RLHF
                    [
                        \citet{lee2023aligning}{,} ReFL ~\citep{xu2023imagereward}{,} DPOK ~\citep{fan2023reinforcement}{,}\\
                        DDPO ~\citep{blacktraining}{,} PRDP ~\citep{deng2024prdp}{,} DRaFT ~\citep{clark2023directly}{,}  \\
                        AlignProp ~\citep{prabhudesai2023aligning}{,} DRTune ~\citep{wu2024deep} , leaf, text width=18em
                    ]
                ]
                [
                    DPO
                    [
                        Diffusion-DPO ~\citep{wallace2024diffusion}{,} D3PO ~\citep{yang2024using}{,} \\
                        Dense Reward \citep{yang2024dense}{,} SPO \citep{liang2024step}{,}\\
                        Diffusion-KTO \citep{li2024aligning}
                        , leaf, text width=18em
                    ]
                ]
            ]
            [
                Test-time Alignment
                [
                    Implicit Guidance
                    [
                        Prompt Optimization, leaf, text width=12em
                    ]
                    [
                        Attention Control, leaf, text width=12em
                    ]
                    [
                        Initial Noise Optimization, leaf, text width=12em
                    ]
                ]
                [
                    Explicit Guidance
                    [
                        Reward-based \\ Input Optimization, leaf, text width=12em
                    ]
                    [
                        Reward-Guided \\ Decoding and Sampling, leaf, text width=12em
                    ]
                ]
            ]
            [
                Alignment Beyond \\ T2I Diffusion Models
                [
                    Video Generation~\citep{prabhudesai2024video,yuan2024instructvideo,wang2024lift,xu2024visionreward,liu2025videodpo,zhang2024onlinevpo,jiang2025huvidpo,huang2025flowdpo,lee2024videorepair,oshima2025inference}{,} \\
                    Image Editing ~\citep{Zhang_2024_CVPR}{,} 3D Generation ~\citep{ye2024dreamreward,zhou2025dreamdpo,ignatyev2025ad}{,} \\
                    Molecule Generation ~\citep{gu2024aligning}{,} Decision Making ~\citep{dong2024aligndiff}{,} \\
                    Audio Generation~\citep{majumder2024tango}{,} Motion Generation~\citep{tan2024sopo}, leaf, text width=30em
                ]
            ]
        ]
        [
            Benchmarks and \\ Evaluation
            [
                Benchmarks
                [
                    Scalar Human \\ Preference Dataset
                    [
                        HPD v1~\citep{wu2023hps}{,} HPD v2~\citep{wu2023hpsv2}{,} \\
                        Pick-a-Pic v1~\citep{kirstain2023pick}{,} ImageRewardDB ~\citep{xu2023imagereward}{,} \\
                        Picsart Image-Social ~\citep{isajanyan2024social}, leaf, text width=18em
                    ]
                ]
                [
                    Multi-dimensional Human \\ Feedback Dataset
                    [
                        MHP~\citep{zhang2024learning}{,}
                        RichHF-18K~\citep{liang2024rich}, leaf, text width=18em
                    ]
                ]
                [
                    AI Feedback Dataset
                    [
                        VisionPrefer~\citep{wu2024multimodal}{,} DreamBench++~\citep{peng2025dreambench}, leaf, text width=18em
                    ]
                ]
            ]
            [
                Evaluation
                [
                    Reward Models
                    [
                        Human Feedback Prediction{,} \\ Correlation with Human Raters, leaf, text width=18em
                    ]
                ]
                [
                    T2I Diffusion Models  
                    [
                        Prompts
                        [
                            MS-COCO~\citep{lin2014microsoft}{,} DiffusionDB~\citep{wang2022diffusiondb}{,} \\
                            MT-Bench~\citep{petsiuk2022human}, leaf, text width=16em
                        ]
                    ]
                    [
                        Image Quality
                        [
                            IS~\citep{salimans2016improved}{,} FID~\citep{heusel2017gans}, leaf, text width=16em
                        ]
                    ]
                    [
                        Human \\ Preference \\ Evaluation
                        [
                            CLIP~\citep{radford2021learning}{,} Aesthetic~\citep{schuhmann2022laion}{,} \\
                            HPS v1~\citep{wu2023hps}{,} HPS v2~\citep{wu2023hpsv2}{,} \\
                            PickScore~\citep{kirstain2023pick}{,} ImageReward ~\citep{xu2023imagereward}{,} \\
                            MPS ~\citep{zhang2024learning}{,}  VP-Score~\citep{wu2024multimodal}{,} \\
                            Social Reward ~\citep{isajanyan2024social},leaf, text width=16em
                        ]
                    ]
                    [
                        Fine-grained \\ Evaluation
                        [
                            DALL-Eval~\citep{cho2023dall}{,} GenEval~\citep{ghosh2023geneval}{,} \\ VPEval~\citep{cho2023visual}{,} HEIM~\citep{lee2023holistic}{,} \\
                            LLMScore~\citep{lu2023llmscore}{,} Style~\citep{somepalli2024measuring}, leaf, text width=16em
                        ]
                    ]
                ]
            ]
        ]
    ]
\end{forest}
}
\caption{The framework of this survey in human alignment of diffusion models and beyond.}
\Description{The main body of this survey is divided into three parts: fundamentals of human alignment, human alignment of diffusion models, as well as benchmark and evaluation.}
\label{fig:framework}
\end{figure*}

In this work, we provide a comprehensive review of the alignment of diffusion models to assist researchers and practitioners in understanding how to align these models with human intentions and preferences. A
\href{https://github.com/xie-lab-ml/awesome-alignment-of-diffusion-models}{literature
list} is made publicly available at GitHub. \cref{fig:framework} illustrates the main framework of this survey. \cref{sec:diffusion} introduces recent advancements in diffusion models, particularly those incorporating alignment technologies. \cref{sec:alignment} explores fundamental alignment techniques and related challenges in human alignment. \cref{sec:align_diffusion} focuses on alignment techniques specific to diffusion models. \cref{sec:eval} reviews benchmarks and evaluation metrics for assessing human alignment of diffusion models. \cref{sec:future} outlines future research directions. \cref{sec:conclusion} concludes our work, summarizing the key findings and their implications for both researchers and practitioners. Our survey provides a thorough understanding of the alignment of diffusion models, identifies research gaps, and informs the development of next-generation models, driving future advancements in the field.

\section{An Overview of Diffusion Models}
\label{sec:diffusion} In this section, we briefly outline recent advancements in diffusion models and elucidate the role that human alignment plays in guiding their development.

\begin{figure}[t]
    \centering
    \includegraphics[width=\textwidth]{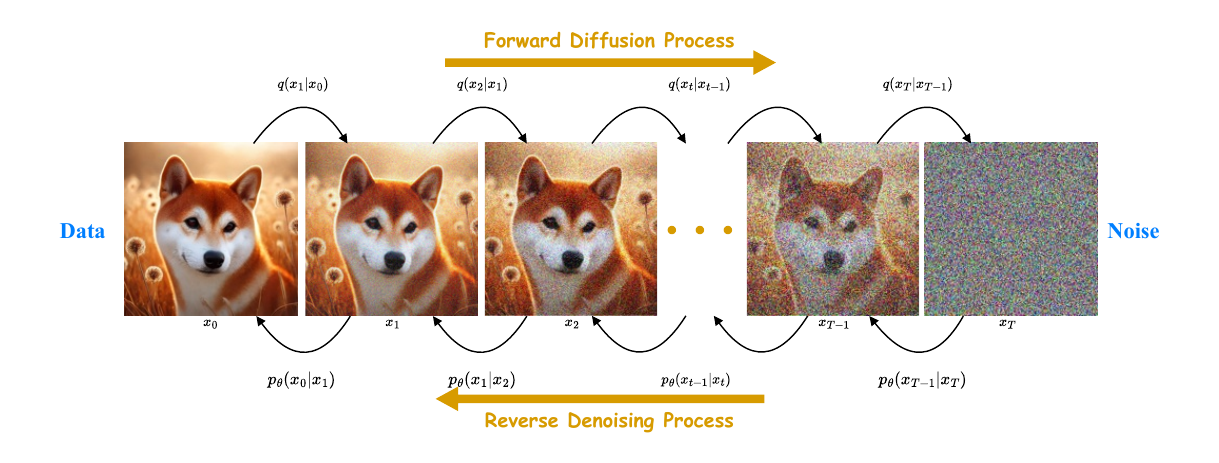}
    \caption{Diffusion models consist of two key processes: a forward diffusion process
    with a transition kernel $q(x_{t}|x_{t-1})$, where noise is gradually added
    to a data sample, and a reverse denoising process with a learnable transition
    kernel $p_{\theta}(x_{t-1}|x_{t})$, where the model learns to denoise
    Gaussian noise to reconstruct the original data sample.}
    \Description{Diffusion models smoothly perturb data by adding noise, then reverse this process to generate new data from noise.}
    \label{fig:diffusion_intuition}
\end{figure}

Decades ago, diffusion process or Langevin diffusion, originated from statistical physics~\citep{bochner1949diffusion,sekimoto1998langevin,gillespie2000chemical}, were first introduced in machine learning not for generative modeling but mainly for parameter inference~\citep{archambeau2007variational} and analyzing optimization dynamics~\citep{xu2018global,cheng2020stochastic,xie2021diffusion}.
{Diffusion models, pioneered by \citet{sohl2015deep} and significantly advanced by \citet{ho2020denoising} with Denoising Diffusion Probabilistic Models (DDPMs), operate through a two-stage process: a forward noising stage that gradually adds noise to data, and a reverse denoising stage that learns to reconstruct data from noise, as illustrated in \cref{fig:diffusion_intuition}. This iterative refinement allows for the generation of high-quality samples. Variations like Denoising Diffusion Implicit Models (DDIMs)~\citep{song2020denoising} further improved sampling efficiency. For detailed reviews, see \citet{yang2023diffusion,cao2024survey}.}

A key application area for diffusion models is T2I synthesis. Early T2I models like GLIDE~\citep{nichol2021glide} adapted diffusion processes for this task but operated in the pixel space, leading to high computational costs~\citep{saharia2022photorealistic}. The advent of Latent Diffusion Models (LDMs)~\citep{rombach2022high}, which perform diffusion in a compressed latent space, marked a significant milestone by reducing computational demands and improving efficiency, making T2I models more practical.

However, even with these advancements, a core challenge emerged: the standard diffusion training objective does not inherently guarantee alignment with nuanced human intentions and preferences. This often results in generated images that, while technically sound, may not fully meet user expectations or desired aesthetics~\citep{podell2023sdxl, betker2023improving, chen2023pixart}. Recognizing this gap, recent cutting-edge T2I models have explicitly integrated human alignment techniques. For example, Stable Diffusion 3 (SD3)~\citep{esser2024scaling} not only introduced architectural innovations but also crucially incorporated alignment by applying Diffusion Direct Preference Optimization (Diffusion-DPO)~\citep{wallace2024diffusion} to its large base models. This alignment step is pivotal in achieving state-of-the-art performance, surpassing other open models and even proprietary ones like DALLE-3~\citep{betker2023improving} on benchmarks such as GenEval~\citep{ghosh2023geneval}. Similarly, SD3-Turbo~\citep{sauer2024fast}, focuses on efficient high-resolution generation, also leverages DPO-finetuned models in its distillation process, demonstrating significant improvements in human preference evaluations. These developments underscore that human alignment is no longer an afterthought but a central component in advancing the capabilities of diffusion models.

This trend highlights the increasing importance of human feedback and sophisticated alignment methodologies in shaping the next generation of diffusion models, moving beyond mere generation quality to achieve outputs that are more accurate, desirable, and aligned with human values. This survey delves into these critical alignment techniques.
\begin{table}[t]
    \centering
    \caption{The list of symbols.}
    \begin{tabular}{c|l}
        \toprule
        Symbols & Meanings \\
        \midrule
        $\mathcal{L}$ & the loss function for optimization \\
        $c$ & the prompt to LLMs or diffusion models \\
        $\rho$ & the distribution of prompt \\
        $x$ & the response of LLMs or diffusion models \\
        $K$ & the number of candidate responses \\
        $x^w$ & the winning/preferred response in the paired responses \\
        $x^l$ & the losing/dis-preferred response in the paired responses \\
        $p_{\mathrm{BT}}$ & the probability distribution of human preference under the Bradley-Terry model \\
        $p_{\mathrm{PL}}$ & the probability distribution of human preference under the Plackett-Luce model \\
        $L$ & the total number of tokens in the responses for LLMs \\
        $T$ & the total number of denoising steps for diffusion models \\
        $\theta$ & the parameters of LLMs or diffusion models \\
        $q$ & the image data distribution \\
        $p_{\theta}$ & the policy in RL, parameterized by $\theta$, i.e., the LLMs or diffusion models to be aligned \\
        $p_{\mathrm{ref}}$ & the reference policy, which is typically the frozen initial policy \\
        $r_{\phi}(c, x)$ & the reward model output given the input prompt $c$ and response $x$, parameterized by $\phi$ \\
        $\mathcal{D}$ & the pre-collected preference dataset \\
        $D_{\mathrm{KL}}$ & the Kullback–Leibler divergence \\
        $\beta$ & the hyper-parameter, which regularizes the distance between the current and reference policies \\
        \bottomrule
        \end{tabular}
    \label{tab:list_of_symbols}
\end{table}
\section{Fundamentals of Human Alignment}
\label{sec:alignment} In this section, we discuss the fundamentals of human alignment based on the existing literature for aligning LLMs and diffusion models.
Specifically, we summarize the general data forms and preference modeling methods for alignment in \cref{sec:alignment_data}. We outline the general alignment algorithms for human alignment in \cref{sec:optimization}. We discuss key challenges
in human alignment in \cref{sec:relevant_problms}.

\subsection{Preference Data and Modeling}
\label{sec:alignment_data}
\textbf{Preference Data.} In general, preference data consists of three elements: prompts, responses\footnote{We use the term ``responses'' broadly to include human-collected data samples beyond model responses.}, and feedback. \cref{tab:list_of_symbols} shows the mathematical notations in this work.

{Preference data are typically composed of prompts, responses, and feedback. Prompts, whether human-provided or AI-generated, are diverse inputs used to elicit model responses. Responses can be generated on-policy (from current model) or off-policy (from external or older models), presenting a trade-off between relevance and collection efficiency~\citep{sutton2020reinforcement}.}

\noindent
\textbf{Preference Modeling.} The most common feedback form is a pairwise preference between a preferred response $x^w$ and a dis-preferred response $x^l$, given a prompt $c$. This is often modeled using a reward model $r_{\phi}$ trained under the Bradley-Terry (BT) framework~\citep{bradley1952rank}. Specifically, the reward model $r$ parameterized with $\phi$ takes in a prompt $c$ and a response $x$, outputting a scalar reward $r_{\phi}(c, x)$. The BT model assumes that the human preference probability $p_{\mathrm{BT}}$ can be expressed as:
\begin{equation}
  \label{eq:BT}p_{\mathrm{BT}}(x^{w} > x^{l} | c) = \frac{\exp(r^{*}(c, x^{w}))}{\exp(r^{*}(c,x^{w})) + \exp(r^{*}(c, x^{l})))}= \sigma(r^{*}(c, x^{w}) - r^{*}(c, x^{l})),
\end{equation}
where $r^{*}$ represents the optimal reward model that $r_{\phi}$ approximates, and $\sigma(x) = 1/(1+\exp(-x))$ is the logistic function. The model is optimized with a loss function that maximizes the log-probability of the preferred sample having a higher reward score:
\begin{equation}
  \label{eq:RM-BT}
 \mathcal{L}_{\mathrm{RM-BT}}(\phi) = - \mathbb{E}_{(c, x^w, x^l) \sim \mathcal{D}}\left[\log (\sigma(r_{\phi}(c, x^{w}) - r_{\phi}(c, x^{l}))) \right],
\end{equation}
where $(c, x^{w}, x^{l}) \sim \mathcal{D}$ denotes the sampling of prompt $c$, the
preferred response $x^{w}$, and the dis-preferred response $x^{l}$ from the collected dataset $\mathcal{D}$ labeled by humans or AI. In essence, \cref{eq:RM-BT} represents a cross-entropy loss where pairwise comparisons are treated as labels, with $x^{w}$ labeled as 1 and $x^{l}$ as 0. The term $\sigma(r_{\phi}(c, x^{w}) - r_{\phi}(c, x^{l}))$ represents the probability that response $x^{w}$ will be preferred over response $x^{l}$ by a human labeler, as modeled by \cref{eq:BT}.

Beyond pairwise comparisons, feedback can be listwise (ranking multiple responses) or binary (a single response is desirable/undesirable) ~\citep{ethayarajh2024kto,richemond2024offline}. Listwise feedback can be modeled by extending the BT model to the Plackett-Luce (PL) model ~\citep{plackett1975analysis, luce1959individual}, which frames alignment as a ranking problem~\citep{yuan2023rrhf, song2024preference, liu2024lipo}.  Specifically, the PL model stipulates that when presented with a set of possible choices, people prefer each choice with a probability proportional to the value of some underlying reward function. In our context, the policy $p$ is given a prompt $c$ and produces a set of $K$ responses $(x_{1}, x_{2}, \dots, x_{K}) \sim p(x|c)$. A human then ranks these responses, yielding a permutation $\tau: [K] \rightarrow [K]$, where $[K]= \{1, 2, \dots, K\}$ indexes the responses and $\tau(i)=j$ indicates that the response $x_j$ is ranked at position $i$. The PL model assumes that the human preference ranking probability $p_{\mathrm{PL}}$ can be formulated as:
\begin{equation}
\label{eq:PL}
p_{\mathrm{PL}}(\tau |x_{1}, x_{2}, \dots, x_{K} , c) = \prod_{k=1}^{K} \frac{\exp(r^{*}(c, x_{\tau(k)}))}{\Sigma_{j=k}^{K} \exp(r^{*}(c, x_{\tau(j)}))}.
\end{equation}
Notably, when $K=2$, \cref{eq:PL} reduces to the BT model in \cref{eq:BT}. The loss function for training the reward model on listwise feedback under the PL model typically uses a maximum likelihood estimation (MLE) ranking loss~\citep{xia2008listwise,rafailov2023direct,liu2024lipo}:
\begin{equation}
\label{eq:RM-PL}
\mathcal{L}_{\mathrm{RM-PL}}(\phi) = - \mathbb{E}_{c, x_1, x_2, \dots, x_K, \tau}\left[\log \prod_{k=1}^{K} \frac{\exp(r_{\phi}(c, x_{\tau(k)}))}{\Sigma_{j=k}^{K} \exp(r_{\phi}(c, x_{\tau(j)}))}\right].
\end{equation}

\noindent
{\textbf{Feedback Sources and Timing.} While traditionally sourced from humans, the cost and effort of annotation have motivated the use of AI-generated feedback, with powerful models like GPT-4 being used as annotators~\citep{cui2023ultrafeedback, wu2024multimodal, pmlr-v235-stephan24a}. Furthermore, feedback can be collected offline from a static dataset or online during training, which distinguishes between offline and online alignment settings~\citep{levine2020offline, dong2024rlhf, tang2024understanding}.}

\subsection{Alignment Algorithms}
\label{sec:optimization}
In this subsection, we introduce general alignment algorithms.

\subsubsection{Reinforcement Learning from Human Feedback}
\label{sec:align_with_reward} 
Alignment with human preferences is typically achieved through RLHF, which first trains an explicit reward model to reflect human preferences and then applies RL methods to optimize a policy toward maximizing the reward provided by the reward model~\citep{christiano2017deep}. RLHF was successfully applied by \citet{ouyang2022training} to fine-tune instruction-following LLMs, leading to the development of the widely-used ChatGPT.

Specifically, the policy $p_{\theta}$ is fine-tuned to maximize the reward $r_{\phi}(c,x)$ while being regularized by the KL divergence from an initial reference policy $p_{\mathrm{ref}}$:
\begin{equation}
    \label{eq:RLHF}\max_{p_{\theta}}\mathbb{E}_{c \sim \rho, x \sim p_{\theta}(x|c)}
    \left[r_{\phi}(c,x) - \beta D_{\mathrm{KL}}(p_{\theta}(x|c)||p_{\mathrm{ref}}
    (x|c)) \right],
\end{equation}
where $\beta$ controls the strength of the KL regularization term~\citep{stiennon2020learning}.

{Proximal Policy Optimization (PPO)~\citep{schulman2017proximal} is the predominant RL algorithm for this task, but it is computationally expensive and notoriously difficult to tune~\citep{ouyang2022training, engstrom2020implementation}. This has motivated simpler alternatives, such as basic REINFORCE-style optimization~\citep{ahmadian2024back, sutton2020reinforcement} or iterative fine-tuning methods that bypass traditional RL entirely by using the reward model to filter or rank data for supervised fine-tuning~\citep{dong2023raft, yuan2023rrhf, liu2024statistical}.}

\subsubsection{Direct Preference Optimization}
\label{sec:align_without_reward}
{DPO offers a simpler paradigm by bypassing explicit reward model training and optimizing the policy directly on preference data~\citep{rafailov2023direct}. By re-parameterizing the RLHF objective (\cref{eq:RLHF}) in terms of the optimal policy, DPO derives a direct loss on pairwise preferences:}
\begin{equation}
    \label{eq:DPO}\mathcal{L}_{\mathrm{DPO}}(p_{\theta};p_{\mathrm{ref}}) = -\mathbb{E}
    _{(c, x^w, x^l) \sim \mathcal{D}}\left[ \log \sigma\left(\beta \log \frac{p_{\theta}(x^{w}|c)}{p_{\mathrm{ref}}(x^{w}|c)}
    - \beta \log \frac{p_{\theta}(x^{l}|c)}{p_{\mathrm{ref}}(x^{l}|c)}\right)\right
    ].
\end{equation}

{Several variants have been proposed to address DPO's limitations, including its tendency to overfit to the offline preference dataset, its sensitivity to hyperparameters, the need for a separate reference model, and its restriction to pairwise preference data. Identity Preference Optimization (IPO)~\citep{azar2024general} modifies the objective with a robust regularization term to mitigate overfitting to the preference dataset. Odds Ratio Preference Optimization (ORPO)~\citep{hong2024orpo} eliminates the need for a separate reference model and combines standard supervised fine-tuning on preferred responses with preference alignment. Other approaches use different feedback formats entirely. Kahneman-Tversky Optimization (KTO)~\citep{ethayarajh2024kto} requires only binary feedback (desirable/undesirable) instead of pairs, leveraging principles from human decision-making theory. Preference Ranking Optimization (PRO) ~\citep{song2024preference} and Listwise Preference Optimization (LiPO) framework ~\citep{liu2024lipo} extend the paradigm to leverage listwise feedback, where multiple responses are ranked.}

\subsection{Challenges of Human Alignment}
\label{sec:relevant_problms} 
{In this subsection, we synthesize key challenges in human alignment.}

\noindent
\textbf{Alignment with AI Feedback.} {Using AI-generated feedback, or Reinforcement Learning from AI Feedback (RLAIF), is a prominent strategy to circumvent costly human annotations~\citep{bai2022constitutional, lee2024rlaif}. This involves models improving themselves via self-generated rewards~\citep{yuan2024self, li2023self} or using powerful proxy models (e.g., LLMs) as annotators~\citep{dubois2023alpacafarm, wu2024multimodal}. However, this introduces a core trade-off between cost-effectiveness and the risks of AI feedback, namely inheriting biases, lacking diversity, and potential model collapse from recursively training on synthetic data~\citep{shumailov2024ai}.}

\noindent
\textbf{Diverse and Changing Human Preferences.} {A key challenge is modeling the diverse, dynamic, and often conflicting nature of human preferences~\citep{carroll2024ai}. Current research addresses this by learning distributions over preferences~\citep{chakraborty2024maxmin}, developing pluralistic frameworks~\citep{sorensen2024roadmap, conitzer2024social}, and using multi-objective alignment for conflicting goals~\citep{rame2023rewarded, pmlr-v235-yang24q}. While these methods promote fairness~\citep{shen2024finetuning}, they also increase algorithmic complexity and introduce the difficulty of balancing competing values.}

\noindent
\textbf{Distributional Shift.} {Reliance on static, offline preference data creates a distributional shift between the training data and the evolving policy, a known challenge in offline RL~\citep{levine2020offline} that leads to over-optimization and reward hacking~\citep{gao2023scaling, skalse2022defining}. KL regularization, a common mitigation, can be overly restrictive and limit model improvement. Future work could draw from fields like OOD generalization~\citep{liu2021towards}, causality~\citep{Sch_lkopf_2022}, and uncertainty estimation~\citep{gal2016dropout,NIPS2017_2650d608} to enhance robustness.} 

\noindent
\textbf{Efficiency of Alignment.} {Improving alignment efficiency is pursued via data-centric and algorithmic approaches. Data-centric methods focus on achieving strong performance with less data, such as by curating small, high-quality instruction sets~\citep{zhou2023lima, liu2024what}. Algorithmic innovations aim to reduce computational overhead through techniques like linear alignment~\citep{gao2024linear} and feedback-efficient exploration~\citep{uehara2024feedback}. The primary challenge is to enhance efficiency without sacrificing alignment quality or introducing data selection biases. Future work may explore dataset distillation~\citep{wang2018dataset,yu2023dataset}, parameter-efficient fine-tuning~\citep{peft}, and inference-time scaling~\citep{snell2024scaling}.}

\noindent
\textbf{Alignment with Rich Rewards.} {A single, terminal reward is often sparse and overlooks the sequential generation process, causing optimization instability~\citep{JMLR:v22:19-736, snell2023offline}. To address this, research explores richer, denser reward structures, such as step-wise preferences for diffusion models~\citep{liang2024step, yang2024dense} and token-level feedback for LLMs~\citep{chan2024dense, zeng2024token}. However, aligning with richer rewards inevitably increases both computational and algorithmic complexities, necessitating further research to address potential scalability issues and to understand how to best design and utilize these complex reward structures.}

\noindent
\textbf{Understanding of Alignment.} Research is ongoing to understand the mechanisms, theoretical properties, and limitations of alignment. {Comparative analyses examine the trade-offs between dominant paradigms like RLHF and DPO, exploring differences in their optimization behavior and performance~\citep{pmlr-v235-ji24c, pmlr-v235-xu24h}. Theoretical studies are also emerging to formalize the learning dynamics of alignment~\citep{pmlr-v235-im24a, pmlr-v235-xiong24a} and analyze fundamental model-level flaws, such as the DPO loss function's potential lack of a unique MLE~\citep{hong2024energy}. Finally, trustworthiness remains a major concern, with research highlighting model vulnerabilities to jailbreak attacks~\citep{pmlr-v235-wolf24a, yi2024jailbreak}, the brittleness of safety alignment~\citep{qi2024finetuning, pmlr-v235-wei24f}, and the negative impact of noisy preference data~\citep{pmlr-v235-ray-chowdhury24a}. Collectively, these studies underscore the gap between current methods and the goal of robust, understandable, and truly aligned AI systems.}

\section{Human Alignment Techniques of Diffusion Models}
\label{sec:align_diffusion} In this section, we first introduce training-based human alignment techniques of diffusion models, including RLHF and DPO in \cref{sec:align_rlhf_diffusion} and \cref{sec:align_dpo_diffusion}, respectively. We then review test-time alignment techniques in \cref{sec:training_free_alignment}. Furthermore, we review studies related to alignment beyond T2I diffusion models in \cref{sec:align_beyond}. Finally, we discuss challenges of diffusion alignment in \cref{sec:align_diffusion_challenges}.

RLHF and DPO are two very classic training-based techniques for aligning AI models with human preferences. However, when applied to diffusion models, these methods encounter significant challenges due to the step-by-step training and sampling nature of diffusion models. Specifically, aligning diffusion models with preference optimization requires sampling across all possible diffusion trajectories leading to $x_{0}$, which is intractable in practice. While the LLM response is treated as a single output, diffusion models' multiple latent image representations of each step need to be calculated and stored, leading to unreasonable high memory consumption and low computation efficiency. This makes these methods impractical for large-scale diffusion models. To address the high computational overhead associated with adapting alignment techniques to diffusion models, researchers often formulate the denoising process as a multi-step Markov decision process (MDP). The proposed diffusion alignment methods need to directly optimize the expected reward of an image output or update the policy based on human preferences to approximately perform policy gradient guided by a reward model. This formulation enables parameter updates at each step of the denoising process based on human preferences, thereby circumventing the significant computational costs. \cref{tab:alignment_comparison} highlights the core differences among RLHF, DPO and test-time alignment techniques for diffusion models.

\begin{table}[thbp]
\caption{Comparison of Human Alignment Paradigms for Diffusion Models.}
\label{tab:alignment_comparison}
\centering
\begin{tabular}{@{}p{1.5cm}p{2.3cm}p{2.4cm}p{2.6cm}p{2.5cm}p{2.6cm}@{}}
\toprule
\textbf{Paradigm} & \textbf{Compute Cost} & \textbf{Feedback / Reward} & \textbf{Scalability} & \textbf{Key Limitations} & \textbf{Best Use Cases} \\
\midrule
\textbf{RLHF} 
& \textbf{High}: multi-step rollouts, trajectory storage, RL optimization 
& \textbf{Explicit}: scalar reward via learned or heuristic reward models 
& \textbf{Low}: limited by annotation cost and unstable RL training 
& High variance, training instability, reward hacking, memory intensive 
& When rich, well-defined rewards are easier to learn than the policy itself \\
\midrule
\textbf{DPO} 
& \textbf{Moderate}: avoids RL loops and explicit reward model training 
& \textbf{Implicit}: relative preference via log-likelihood ratios 
& \textbf{Moderate}: simpler pipeline but depends on high-quality preference pairs 
& Sensitive to distribution shift; limited robustness outside preference data 
& When preferences are available but reward modeling is unreliable or costly \\
\midrule
\textbf{Test-Time Alignment} 
& \textbf{Low--Moderate}: added inference-time optimization, no retraining 
& \textbf{Hybrid}: heuristics or external (differentiable / black-box) rewards 
& \textbf{High}: model-agnostic and training-free 
& Increased inference latency; local or proxy-driven failures possible 
& Lightweight, on-the-fly, or personalized alignment without model updates \\
\bottomrule
\end{tabular}
\end{table}

\subsection{Reinforcement Learning from Human Feedback of Diffusion Models}
\label{sec:align_rlhf_diffusion} 
\begin{figure}[t]
    \centering
    \includegraphics[width=\textwidth]{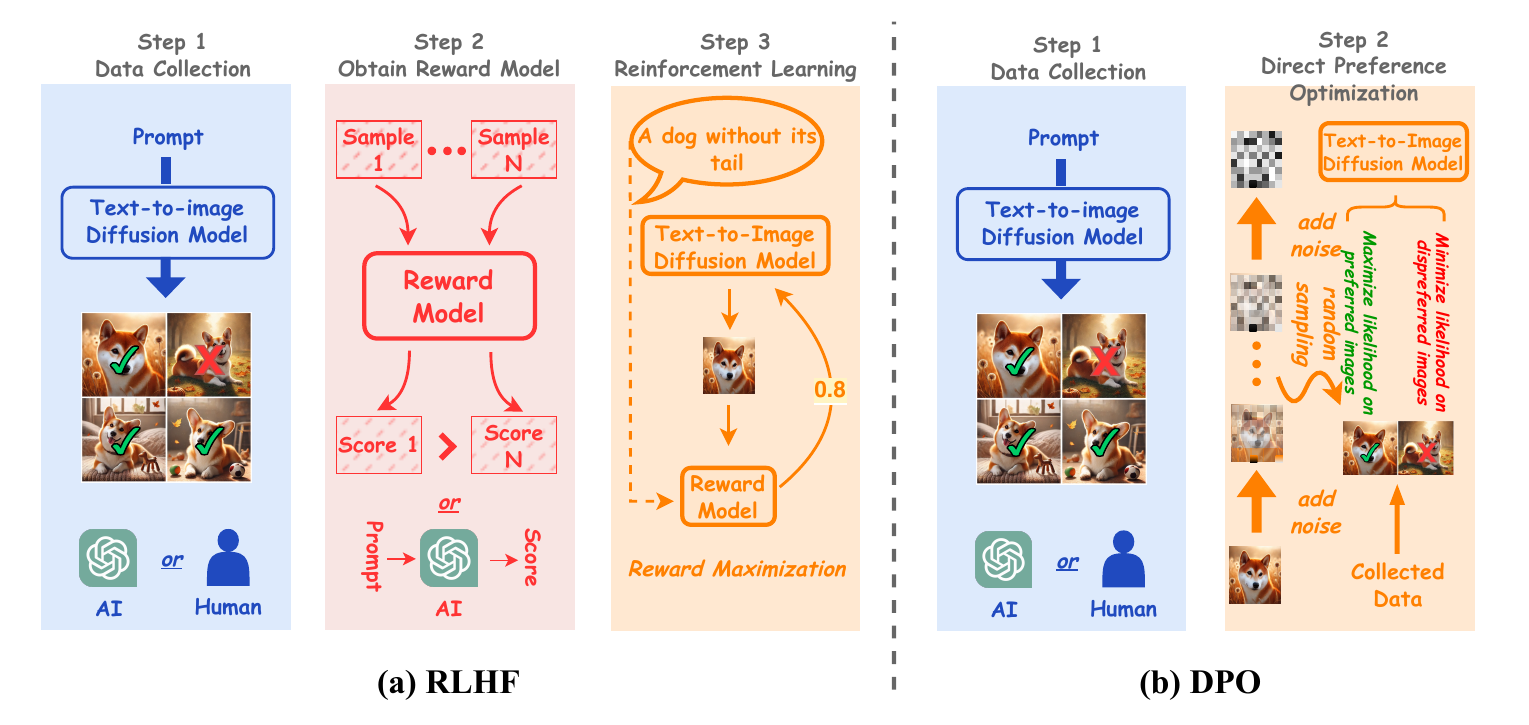}
    \caption{The overview of RLHF and DPO of diffusion models.}
    \Description{RLHF in T2I generation typically involves three progressive stages: data collection, developing a reward model, and reinforcement learning. DPO in T2I generation typically involves two stages: data collection and direct preference optimization that aligns models with human preferences by directly optimizing them on human preference data. }
    \label{fig:sec4_rlhf_dpo}
\end{figure}

In this subsection, we present the RLHF paradigm and its extension for diffusion alignment.
As shown in \cref{fig:sec4_rlhf_dpo} (a), RLHF typically involves three progressive stages: data collection, developing a reward model, and reinforcement learning. In the data collection stage, preferences of prompt-response pairs (e.g., text-image pairs for T2I diffusion models) are gathered from humans or AI. In the second stage, RLHF develops a reward model $r_{\phi}(c, x)$, either through training or prompt engineering~\citep{lu2023llmscore}. The trained reward model for diffusion models is often instantiated as a VLM such as CLIP~\citep{radford2021learning} or BLIP~\citep{li2022blip} and typically trained with \cref{eq:RM-BT}~\citep{xu2023imagereward, kirstain2023pick, wu2023hpsv2} on $\mathcal{D}$ to model human preferences. Finally, RLHF optimizes the diffusion model $p_{\theta}(x_{0}|c)$ to maximize the reward $r_{\phi}(c, x_{0})$ given the prompt distribution $c \sim \rho$ (ignoring the regularization
term):
\begin{equation}
    \label{eq:diffusion_RLHF_0}\min_{\theta}\mathbb{E}_{c \sim \rho, x_0 \sim p_\theta(x_0|c)}
    \left[-r_{\phi}(c, x_{0})\right].
\end{equation}
{There are several classical and emerging approaches to approximately optimizing the objective in \cref{eq:diffusion_RLHF_0}. They can be broadly categorized into reward-weighted fine-tuning, RL fine-tuning, and direct reward fine-tuning, alongside newer paradigms.}

\noindent
\textbf{Reward-weighted Fine-tuning.} \citet{lee2023aligning} proposed to align
diffusion models with human feedback with a reward-weighted likelihood
maximization objective:
\begin{equation}
    \label{eq:lee et al.}\min_{\theta}\mathbb{E}_{(c, x_0) \sim \mathcal{D_{\mathrm{model}}}}
    \left[-r_{\phi}(c, x_{0}) \log p_{\theta}(x_{0}|c)\right] + \beta \mathbb{E}_{(c,
    x_0) \sim \mathcal{D_{\mathrm{pre-training}}}}\left[- \log p_{\theta}(x_{0}|c
    )\right],
\end{equation}
where $(c, x_{0}) \sim \mathcal{D_{\mathrm{model}}}$ is the model-generated dataset
by diffusion models on the tested text prompts, and
$\mathcal{D_{\mathrm{pre-training}}}$ is the pre-training dataset. The first term
in \cref{eq:lee et al.} minimizes the reward-weighted negative log-likelihood (NLL)
on $\mathcal{D_{\mathrm{model}}}$ to improve the image-text alignment of the model.
The second term in \cref{eq:lee et al.} minimizes the pre-training loss to
mitigate overfitting to $\mathcal{D_{\mathrm{model}}}$. \citet{blacktraining}
pointed out that \cref{eq:lee et al.} can be performed for multiple rounds of
alternating sampling and training to make it into an online RL method by
replacing $(c, x_{0}) \sim \mathcal{D_{\mathrm{model}}}$ with
$c \sim \rho, x_{0} \sim p_{\theta}(x_{0}|c)$. They referred to this general class
of reward-weighted methods as reward-weighted regression (RWR), and considered
two weighting schemes: 1) a standard one that uses exponentiated rewards to
ensure nonnegativity, $w_{\mathrm{RWR}}(c,x_{0}) = \frac{1}{\mathcal{Z}_{RWR}}\exp
(\gamma r_{\phi}(c, x_{0}))$, where $\gamma$ is an inverse temperature and
$\mathcal{Z}_{RWR}$ is a normalization constant; and 2) a simplified one that
uses binary weights
$w_{\mathrm{sparse}}(c,x_{0}) = \mathbb{I}\left[r_{\phi}(c, x_{0}) \geq C \right]$,
where $C$ is a reward threshold determining which samples are used for training and
$\mathbb{I}$ is the indicator function. Notably, from the RL literature, a weighted
log-likelihood objective by $w_{\mathrm{RWR}}$ is known to approximately solve
\cref{eq:diffusion_RLHF_0} subject to a KL divergence constraint on
$p_{\theta}(x_{0}|c)$~\citep{nair2020awac}.

\noindent
\textbf{RL Fine-tuning.} Reward-weighted fine-tuning relies on an approximate log-likelihood because it ignores the sequential nature of the diffusion denoising process, only using the final samples $x_{0}$. To address this, the denoising process is treated as a multi-step decision-making problem~\citep{fan2023reinforcement, blacktraining, uehara2024understanding}, using exact likelihoods at each denoising step instead of the approximate likelihoods from the full denoising process. This allows us to directly optimize \cref{eq:diffusion_RLHF_0} using policy gradient algorithms. \citet{blacktraining} proposed denoising diffusion policy optimization (DDPO) to maximize rewards from various reward models, including image compressibility, aesthetic quality, and image-prompt alignment. They demonstrated that DDPO is more effective than reward-weighted likelihood approaches. DDPO has two variants. One uses REINFORCE~\citep{williams1992simple,mohamed2020monte}, a score function policy gradient estimator:
\begin{equation}
    \label{eq:DDPO_SF}\mathbb{E}_{c \sim \rho, x_{0:T} \sim p_\theta(x_{0:T}|c)}\left
    [ -r_{\phi}(c, x_{0}) \Sigma_{t=1}^{T} \nabla_{\theta}\log p_{\theta}(x_{t-1}
    |x_{t},c) \right].
\end{equation}
Another variant uses an importance sampling estimator to reuse old trajectories (i.e., prompt-image pairs) {with a PPO-style clipping objective~\citep{schulman2017proximal} to stabilize training}:
\begin{equation}
    \label{eq:DDPO_clip}
    \mathbb{E}_{c \sim \rho, x_{0:T} \sim p_{\mathrm{ref}}(x_{0:T}|c)} \left[ -r_{\phi}(c, x_0) \Sigma_{t=1}^T \mathrm{clip} \left( \frac{p_{\theta}(x_{t-1}|x_t,c)}{p_{\mathrm{ref}}(x_{t-1}|x_t,c)} \nabla_{\theta} \log p_{\theta}(x_{t-1}|x_t,c), 1-\epsilon, 1+\epsilon \right)\right],
\end{equation}
where $\epsilon$ is the clip hyperparameter.

\citet{fan2023reinforcement} introduced Diffusion Policy Optimization with KL regularization (DPOK), an online RL fine-tuning algorithm that maximizes the ImageReward score with KL regularization. Compared to DDPO, DPOK~\citep{fan2023reinforcement} further employs KL regularization to \cref{eq:diffusion_RLHF_0}, resulting in the
objective:
\begin{equation}
    \min_{\theta}\mathbb{E}_{c \sim \rho, x_0 \sim p_\theta(x_0|c)}\left[-r_{\phi}
    (c, x_{0}) + \beta D_{\mathrm{KL}}(p_{\theta}(x_{0}|c)||p_{\mathrm{ref}}(x_{0}
    |c)) \right].
\end{equation}
DPOK then utilizes an upper bound of the KL-term to derive the following objective for regularized training:
\begin{equation}
    \label{eq:DPOK}\min_{\theta}\mathbb{E}_{c \sim \rho, x_{0:T} \sim p_\theta(x_{0:T}|c)}
    \left[-r_{\phi}(c, x_{0}) \right] + \beta \Sigma_{t=1}^{T} \mathbb{E}_{x_t
    \sim p_\theta(x_t|c)}\left[ D_{\mathrm{KL}}(p_{\theta}(x_{t-1}|x_{t},c)||p_{\mathrm{ref}}
    (x_{t-1}|x_{t},c)\right].
\end{equation}
{However, policy gradient methods like DDPO are known for their high variance and instability, especially in large-scale settings. To address this, \citet{deng2024prdp} proposed Proximal Reward Difference Prediction (PRDP), which reframes the RL objective as a more stable, supervised reward difference prediction task. Instead of directly estimating policy gradients, PRDP trains the model to predict the reward difference between two generated images, proving that a model which perfectly predicts this difference effectively maximizes the original RL objective.}

\noindent
\textbf{Direct Reward Fine-tuning.} RL fine-tuning methods are flexible because they do not require differentiable rewards. However, many reward models are differentiable, such as ImageReward, PickScore~\citep{kirstain2023pick}, and HPSv2~\citep{wu2023hpsv2}, providing analytic gradients. In such cases, using RL can discard valuable information from the reward model. To address this, end-to-end backpropagation from reward gradients to the diffusion model parameters has been
proposed to solve \cref{eq:diffusion_RLHF_0}. Nevertheless, updating the diffusion model throughout the entire denoising process is memory-intensive, as storing partial derivatives of all layers and denoising steps is prohibitive. ReFL~\citep{xu2023imagereward} was the first to backpropagate through a differentiable reward model by evaluating the reward on a one-step predicted image $r(c, \hat{x_0})$ from a randomly selected step $t$, thus bypassing the full denoising process. In contrast, Alignment by Backpropagation (AlignProp)~\citep{prabhudesai2023aligning} and Direct Reward Fine-Tuning (DRaFT)~\citep{clark2023directly} evaluate the reward on the final iteratively denoised image $x_{0}$. Techniques such as low-rank adaptation (LoRA)~\citep{hulora}
and gradient checkpointing~\citep{chen2016training} are employed to reduce memory costs. {To address the ``depth-efficiency dilemma'' of backpropagating through many steps, Deep Reward Tuning (DRTune)~\citep{wu2024deep} enables more efficient deep supervision by stopping the gradient of the denoising network's input and training only on a selective subset of steps.} Direct reward fine-tuning avoids the high variance and low sample efficiency inherent in RL fine-tuning, thus improving training efficiency. However, fine-tuning with a differentiable reward model introduces a risk of over-optimization, potentially resulting in high-reward but lower-quality images. To mitigate this, DRaFT~\citep{clark2023directly} explores methods such as LoRA scaling, early stopping, and KL regularization, with LoRA scaling found to be the most effective in reducing reward overfitting.

\noindent
{\textbf{Advanced Fine-tuning Paradigms and Strategies.}
Beyond these foundational approaches, recent works have introduced more sophisticated paradigms to tackle key challenges like reward over-optimization, diversity collapse, and the alignment of specialized, fast models.

\noindent
\textit{Tackling Reward Over-optimization}: A primary failure mode is ``reward hacking'', where the model maximizes the reward metric at the expense of true image quality. \citet{zhang2024confronting} proposed Temporal Diffusion Policy Optimization (TDPO-R), which provides reward supervision at each denoising step rather than just on the final image. This temporal inductive bias, combined with a ``critic active neuron reset'' mechanism to combat overfitting to early experiences (primacy bias), improves robustness. This aligns with theoretical work framing fine-tuning as entropy-regularized control to prevent reward collapse~\citep{chen2024fine}.

\noindent
\textit{Aligning Fast and Few-Step Models}: As diffusion models become faster through distillation, aligning them presents new challenges. Stepwise Diffusion Policy Optimization (SDPO)~\citep{cheng2024aligning} is designed for few-step models by learning from dense rewards at each step. For ultra-fast (e.g., 1-2 step) models where standard RL fails, LaSRO~\citep{zheng2024reward} learns a differentiable surrogate reward in the latent space, enabling effective gradient-based tuning.

\noindent
\textit{Innovations in Reward Signals and Applications}: The design of the reward function itself is a major area of research. For instance, CoMat~\citep{li2024comat} uses an image-to-text model to provide concept-level rewards, addressing issues of concept ignorance and mismapping. Other works have focused on specialized alignment goals, such as improving long-text alignment~\citep{zhang2024improving}, performing region-aware fine-tuning to fix local flaws~\citep{zhong2025focusnfix}, or using information-theoretic objectives for alignment~\citep{zhang2024information}. \citet{hu2025towards} have demonstrated alignment with only sparse terminal rewards.

\noindent
\textit{Emerging Paradigms}: Other novel approaches are also being explored. Adversarial Diffusion Tuning (ADT)~\citep{huang2025adt} uses an adversarial discriminator to close the gap between the training and inference distributions. To preserve sample diversity, a known issue in reward-driven fine-tuning, \citet{shen2024efficient} proposed using Gradient-Informed GFlowNets ($\nabla$-GFlowNet) to balance reward maximization with exploration.
}

{\textbf{Summary and Outlook} The landscape of fine-tuning diffusion models with RLHF is rapidly evolving from three foundational pillars—reward-weighted regression, policy-gradient RL, and direct reward optimization—to a more diverse set of specialized and robust techniques. While RWR is simple, it offers less precise control. Policy gradient methods like DDPO provide strong optimization capabilities but can suffer from high variance; this is addressed by more stable alternatives like PRDP. Direct reward methods like DRaFT are sample-efficient but risk memory overhead and reward over-optimization, a challenge that methods like DRTune and TDPO-R aim to mitigate. Furthermore, emerging paradigms like adversarial tuning and GFlowNet-based alignment are being developed to address fundamental issues like distribution shift and diversity collapse. A critical consideration in practical applications is feedback efficiency; when reward signals are expensive, online fine-tuning methods like SEIKO~\citep{uehara2024feedback} use uncertainty modeling to minimize queries. The choice of method thus involves a trade-off between implementation complexity, training stability, sample efficiency, and the specific alignment goal, from general aesthetic improvement to targeted concept or regional correction.}

\subsection{Direct Preference Optimization of Diffusion Models}
\label{sec:align_dpo_diffusion}
In this subsection, we present the DPO paradigm
and its extension for diffusion alignment. This paradigm has been successfully adapted to diffusion models, directly optimizing them on human preference data without an explicit reward model as illustrated in \cref{fig:sec4_rlhf_dpo} (b). This approach forms the basis for powerful open-source models like SD3.

\noindent
\textbf{Foundational DPO Methods.}
The core method, Diffusion-DPO~\citep{wallace2024diffusion}, adapts the original DPO objective (\cref{eq:DPO}) to the iterative nature of diffusion models. Specifically, Diffusion-DPO formulated the objective function over the entire diffusion path $x_{0:T}$ as
\begin{eqnarray}
\lefteqn{\mathcal{L}_{\mathrm{Diffusion-DPO}} (p_{\theta};p_{\mathrm{ref}})} \nonumber \\
& = & - \mathbb{E}_{(c, x^w_0, x^l_0) \sim \mathcal{D}} \log \sigma \Biggl(  
\beta \mathbb{E}_{x^w_{1:T} \sim p_{\theta}(x^w_{1:T} | x^w_0, c), x^l_{1:T} \sim p_{\theta}(x^l_{1:T} | x^l_0, c)} \left[\log \frac{p_{\theta}(x^w_{0:T}|c)}{p_{\mathrm{ref}}(x^w_{0:T}|c)} - \log \frac{p_{\theta}(x^l_{0:T}|c)}{p_{\mathrm{ref}}(x^l_{0:T}|c)} \right] \Biggr).
\label{eq:diffusion_DPO_1}
\end{eqnarray}
\cref{eq:diffusion_DPO_1} can be upper bounded~\citep{wallace2024diffusion,yang2024using} using Jensen's inequality and the convexity of function $-\log \sigma$ to push the expectation outside:
\begin{equation}
\label{eq:diffusion_DPO_2}
    \begin{split}
    \mathcal{L}_{\mathrm{Diffusion-DPO}} (p_{\theta};p_{\mathrm{ref}}) \leq - & \mathbb{E}_{(c, x^w_0, x^l_0) \sim \mathcal{D}, t \sim \mathcal{U}(0,T), x^w_{t-1, t} \sim p_{\theta}(x^w_{t-1, t} | x^w_0, c), x^l_{t-1, t} \sim p_{\theta}(x^l_{t-1, t} | x^l_0, c)} \\
    & \log \sigma \Biggl(\beta T \left[\log \frac{p_{\theta}(x^w_{t-1}|x^w_t, c)}{p_{\mathrm{ref}}(x^w_{t-1}|x^w_t, c)} - \log \frac{p_{\theta}(x^l_{t-1}|x^l_t, c)}{p_{\mathrm{ref}}(x^l_{t-1}|x^l_t, c)} \right] \Biggr).
    \end{split}
\end{equation}
Because sampling from the reverse joint distribution $p_{\theta}(x_{t-1, t} | x_0, c)$ is intractable, \citet{wallace2024diffusion} approximates the reverse process $p_{\theta}(x_{1:T} | x_0, c)$ with the forward process $q(x_{1:T} | x_0, c)$. The right-hand side of \cref{eq:diffusion_DPO_2} becomes:
\begin{equation}
\label{eq:diffusion_DPO_3}
    \begin{split}
    \mathcal{L} (p_{\theta};p_{\mathrm{ref}}) = - & \mathbb{E}_{(c, x^w_0, x^l_0) \sim \mathcal{D}, t \sim \mathcal{U}(0,T), x^w_t \sim q(x^w_t | x^w_0, c), x^l_t \sim q(x^l_t | x^l_0, c)}  \\
    & \log \sigma \biggl(- \beta T \biggl(D_{\mathrm{KL}}\left(q(x^w_{t-1}|x^w_{0,t}, c) || p_{\theta}(x^w_{t-1}|x^w_t, c)\right) - D_{\mathrm{KL}}\left(q(x^w_{t-1}|x^w_{0,t}, c) || p_{\mathrm{ref}}(x^w_{t-1}|x^w_t, c)\right)  \\
    & - D_{\mathrm{KL}}\left(q(x^l_{t-1}|x^l_{0,t}, c) || p_{\theta}(x^l_{t-1}|x^l_t, c)\right) + D_{\mathrm{KL}}\left(q(x^l_{t-1}|x^l_{0,t}, c) || p_{\mathrm{ref}}(x^l_{t-1}|x^l_t, c)\right)\biggr)\biggr).
    \end{split}
\end{equation}
This allows the model to learn from preference pairs $(x_0^w, x_0^l)$ by adjusting the denoising process at each step. A key variant, D3PO~\citep{yang2024using}, shares a similar objective but differs primarily in its sampling strategy, using the model's reverse process to obtain noisy latents rather than the forward noising process. Other works have explored alternative optimization perspectives, such as reinterpreting DPO as preference-weighted score matching~\citep{zhu2025dspo}, distribution optimization~\citep{kawata2025direct}, or extending it to handle multiple reward signals~\citep{lee2025calibrated}.

\noindent
{
\textbf{Addressing Temporal Inconsistency: Step-Aware Preference.}
A key limitation of the foundational approach is the assumption that preferences are constant across all denoising steps. This overlooks that different stages of generation influence different aspects of the final image~\citep{hertz2022prompt}. To address this, recent works introduce step-aware or temporally-discounted preferences. This provides more granular feedback throughout the generation process, which has been shown to improve alignment by relaxing the assumption of a static preference signal~\citep{yang2024dense, liang2024step, ren2025refining, lu2025smpo}.

\noindent
\textbf{Enhancements and Extensions of the DPO Framework.}
Further extensions to the DPO framework focus on two main areas: improving the training process and adapting to new contexts. Enhancements to the training process include curriculum-based strategies that use progressively harder examples~\citep{croitoru2025curriculum}, reference-free objectives like MaPO to improve robustness to distribution shifts~\citep{hong2024margin}, combining fine-tuning with test-time sampling~\citep{fu2025chats}, and constructing more visually consistent preference pairs to enhance the training signal~\citep{hu2025dfusion}. In parallel, the framework has been adapted for new contexts, most notably to handle different feedback formats. Diffusion-KTO~\citep{li2024aligning}, for instance, extends the paradigm to use simpler binary feedback (desirable/undesirable) instead of pairwise comparisons.

\noindent
\textbf{Summary and Outlook} The DPO landscape for diffusion models is evolving rapidly. Foundational methods like Diffusion-DPO offer a simple and effective baseline. However, their core assumption of static preferences has motivated the development of more complex but potentially more accurate step-aware alignment techniques. Meanwhile, other research avenues focus on improving training efficiency through better data and learning strategies or expanding the paradigm's applicability to different feedback types and model architectures. The choice of method involves a trade-off between the simplicity of the core framework and the increased granularity and robustness offered by its more advanced extensions.
}

\subsection{Test-Time Alignment of Diffusion Models}
\label{sec:training_free_alignment}
In this subsection, we review methods for aligning diffusion models at test-time without requiring model fine-tuning. This paradigm, also referred to as inference-time alignment \citep{uehara2025inference}, has rapidly evolved from strategies that \textbf{implicitly guide} generation by manipulating initial inputs and internal mechanisms, to more sophisticated approaches that employ \textbf{explicit reward-guided} strategies. In the latter, external preference models (e.g., for aesthetics or semantic consistency) directly steer the sampling process. This evolution offers more powerful and targeted control over model outputs, addressing complex human preferences.

\subsubsection{Implicit Guidance: Optimizing Inputs and Internal Mechanisms}
{Implicit guidance strategies focus on improving alignment without an external reward function, typically by adjusting the initial conditions or internal states of the generation process. These methods are generally computationally lightweight.}

\noindent
\textbf{Prompt Optimization.} Prompt design plays a crucial role in determining generation quality and helping models better understand user intentions~\citep{reynolds2021prompt,zhou2022large}. While manual prompt engineering is common~\citep{liu2022design,oppenlaender2023taxonomy}, it can be labor-intensive. Consequently, recent works have explored optimizing prompts automatically. \citet{wang2023reprompt} developed RePrompt to refine text prompts toward more precise emotional expressions. \citet{hao2023optimizing} introduced Promptist, which adapts user input to model-preferred prompts via RL, where the reward is derived from metrics like CLIP similarity~\citep{radford2021learning} and an aesthetic predictor~\citep{schuhmann2022laion}. \citet{manas2024improving} proposed OPT2I, which leverages an LLM to iteratively revise user prompts, and \citet{mo2024dynamic} introduced an online RL strategy to generate dynamic fine-control prompts.

\noindent
\textbf{Attention Control.}
Attention control has emerged as a crucial technique for improving image-prompt alignment, addressing issues like attribute leakage and missing entities~\citep{feng2022training,rassin2024linguistic}. \citet{hertz2022prompt} first demonstrated that manipulating attention maps can guide the generation process. Building on this, Attend-and-Excite~\citep{chefer2023attend} enhances cross-attention interactions for more semantically aligned content. \citet{wu2024towards} propose a phase-wise attention modulation technique, and \citet{li2023divide} refined these methods for more complex semantic alignments. By dynamically adjusting focus, these mechanisms allow diffusion models to better align their outputs with human expectations~\citep{zheng2023layoutdiffusion,yang2024mastering,hong2023improving}.

\noindent
\textbf{Initial Noise Optimization.}
The reverse diffusion process is highly sensitive to the initial noise~\citep{xu2025good}. To find better noise without a reward model, \citet{qi2024not} introduced the concept of inversion stability. \citet{guo2024initno} introduced Initial Noise Optimization (InitNO), which uses the model's internal attention scores to guide the initial noise towards semantically valid regions.

\subsubsection{Explicit Reward-Guided Strategy: Trajectory Optimization with External Preferences}
{In contrast to implicit methods, explicit reward-guided strategies directly use an external reward function (differentiable or black-box) to modify the sampling trajectory at each step to maximize a preference objective. These methods offer stronger control but typically increase inference complexity.

\noindent
\textbf{Reward-based Input Optimization.}
One approach is to use the reward signal to optimize the initial inputs. For instance, ReNeg \citep{li2025reneg} learns an optimal, universal negative embedding from reward signals. Others directly optimize the initial noise via gradient ascent on a reward function, as seen in Direct Noise Optimization (DNO) \citep{tang2024inference} and ReNO \citep{eyring2024reno}. This concept has been further abstracted into learning a ``noise prompt network'' to generate a tailored ``golden noise'' \citep{zhou2025golden} or framing it as a search problem guided by verifier feedback \citep{ma2025inference}.

\noindent
\textbf{Reward-Guided Decoding and Sampling.}
A more powerful class of methods steers the entire decoding trajectory. Many are derivative-free, enabling the use of black-box reward models. For example, \citet{li2024derivative} propose Soft Value-Based Decoding (SVBD), which uses a soft value function to estimate future rewards. To combat reward over-optimization and maintain diversity, methods based on Sequential Monte Carlo (SMC) \citep{kim2025test} or Feynman-Kac steering \citep{singhal2025general} run and resample multiple generation trajectories. \citet{xie2025dymo} introduced DyMO for multi-objective alignment, while Zigzag Diffusion Sampling (Z-Sampling) \citep{lichen2025zigzag} introduces a self-reflection mechanism. \citet{yeh2025trainingfree} proposed Sampling Demons, a backpropagation-free method that performs stochastic optimization on the sampling distribution. On a theoretical level, \citet{shi2024preference} work towards a more formal connection between terminal preference labels and the generation trajectory.}

\subsubsection{Summary and Outlook}
Test-time alignment techniques provide efficient and flexible ways to improve generation without retraining, evolving from \textbf{implicit guidance} to \textbf{explicit reward-guided} strategies. Implicit methods (e.g., prompt optimization or attention control) guide generation via heuristic or model-intrinsic modifications with minimal inference overhead, but offer limited control precision. In contrast, explicit methods incorporate external preference or reward models to directly optimize the sampling trajectory at inference time, enabling stronger alignment at the cost of higher computation and potential reward hacking.

Overall, the key distinction lies in whether an explicit preference objective is introduced at inference time: implicit methods modify inputs or internal states without an explicit objective, whereas explicit methods optimize generation with respect to an external preference signal. Control-input methods (e.g., spatial masks or sketches) are beyond the scope of this survey, as they specify \emph{what} to generate rather than \emph{how} preference alignment is enforced during sampling; see \citet{11304732} for a comprehensive survey of controllable generation.

For practitioners, the choice of method involves a trade-off between control fidelity and inference efficiency. A promising future direction may involve combining training-based alignment with test-time methods. The former can instill general preferences into the model, while the latter can provide on-the-fly, personalized guidance, leading to more robust and versatile alignment systems.

\subsection{Beyond T2I Diffusion Models}
\label{sec:align_beyond}
{%
In this subsection, we review studies focused on the alignment of non-T2I diffusion models across various generation domains. While pioneering efforts have been made, adapting techniques from T2I diffusion models remains challenging due to domain-specific requirements in preference dataset collection, reward modeling, and the alignment techniques themselves. Each domain thus requires tailored adaptations to achieve effective alignment with human preferences.

\subsubsection{Video Generation}
Aligning video generation models~\citep{10.1145/3487891} with human preferences introduces unique challenges, including immense computational overhead, a scarcity of large-scale video preference datasets, and the complex need to evaluate temporal consistency. Research in this area mirrors the methodological evolution seen in T2I alignment.

\noindent
\paragraph{RLHF-based Alignment.}
Early approaches explored adapting the RLHF framework. For instance, \citet{yuan2024instructvideo} proposed InstructVideo, which reduces fine-tuning costs by using partial DDIM sampling and repurposing image reward models for video. Focusing on more direct reward supervision, VADER~\citep{prabhudesai2024video} and LiFT~\citep{wang2024lift} align models via direct reward fine-tuning. VADER leverages pre-trained reward models for efficiency, while LiFT develops a custom video-specific reward model (LiFT-Critic) trained on a new dataset with textual rationales (LiFT-HRA). The VisionReward framework further advances this by learning fine-grained, multi-dimensional preferences for both images and videos~\citep{xu2024visionreward}. These methods, however, rely heavily on the quality and availability of explicit (or proxy) reward signals.

\noindent
\paragraph{DPO-based Alignment.}
Inspired by its success and simplicity in the image domain, DPO was quickly adapted for video. \citet{liu2025videodpo} pioneered VideoDPO, which uses a comprehensive score to handle both visual quality and semantic alignment. To overcome the static nature of offline datasets, \citet{zhang2024onlinevpo} introduced OnlineVPO, an online DPO algorithm that uses a synthetically-trained video quality assessment (VQA) model for feedback. Other works like HuViDPO~\citep{jiang2025huvidpo} focus on specific niches like human-centric videos, while Flow-DPO~\citep{huang2025flowdpo} demonstrates the paradigm's versatility by applying it to flow-based models. The primary challenge for DPO-based methods is constructing preference pairs that meaningfully capture the temporal nuances of video.

\noindent
\paragraph{Test-time Alignment.}
For lightweight alignment without retraining, test-time methods are emerging. \citet{lee2024videorepair} presented VideoRepair, a model-agnostic framework that identifies and performs localized refinements during inference. Similarly, \citet{oshima2025inference} proposed using a diffusion latent beam search to improve perceptual quality without any model updates. These approaches offer great flexibility but may be limited in addressing complex, global alignment issues.

\subsubsection{Audio and Motion Generation}
Preference alignment is also proving effective for other types of sequential data generation.
In \textbf{audio generation}, \citet{majumder2024tango} introduced Tango 2, which applies diffusion-DPO to fine-tune a text-to-audio model. It uses a synthetic preference dataset where ``loser'' samples feature missing concepts or incorrect temporal ordering. This work highlights DPO's effectiveness in capturing nuanced structural and semantic aspects of audio.
For \textbf{motion generation}, \citet{tan2024sopo} proposed SoPo, a DPO-based method that uses semi-online preference optimization. By combining online and offline data pairs, SoPo addresses the overfitting and sampling bias issues common in standard DPO, leading to higher-quality, human-preferred motions.

\subsubsection{Image Editing}
For the task of instructional visual editing, alignment must consider not only the final quality but also fidelity to user instructions. \citet{Zhang_2024_CVPR} proposed Harnessing Human Feedback for Instructional Visual Editing (HIVE), an RLHF-based approach. It involves collecting feedback on the edited images to learn a reward function, which then guides the fine-tuning of diffusion models to better adhere to human preferences.

\subsubsection{3D Generation}
Aligning text-to-3D models involves challenges beyond aesthetics, such as ensuring multi-view consistency and geometric plausibility.

\noindent
\paragraph{RLHF and DPO-based Alignment.}
\citet{ye2024dreamreward} developed DreamReward, a two-stage RLHF-like process that first trains a 3D-aware reward model (Reward3D) and then uses it to fine-tune the generator. While this introduced the first 3D preference dataset, \citet{zhou2025dreamdpo} demonstrated a more direct path with DreamDPO, which applies DPO to leverage relative rankings, simplifying the need for absolute quality scores. This mirrors the RLHF-to-DPO trend observed in the other domains.

\noindent
\paragraph{Geometric and Semantic Alignment.}
Moving beyond preference scores, \citet{ignatyev2025ad} addressed alignment from a geometric perspective. Their method optimizes for smooth and plausible transitions between generated objects in a latent space, directly enforcing structural consistency. This unique approach enables applications like 3D editing and hybridization, highlighting that 3D alignment is a multi-faceted problem concerning both preference and geometry.

\subsubsection{Specialized Scientific and Control Applications}
Alignment techniques are also applied to specialized domains with functional, rather than purely aesthetic, objectives.
In \textbf{molecule generation}, \citet{gu2024aligning} introduced \textsc{AliDiff}, which aligns diffusion models with desired functional properties like binding affinity via preference optimization. A key challenge here is the accuracies of user-defined reward functions, which may not perfectly model real-world chemical properties.
In \textbf{decision making}, \citet{dong2024aligndiff} used RLHF to guide a planning diffusion model. This allows the model to plan for desired behaviors based on human preferences, demonstrating the potential of alignment for customizing agentic systems.

\subsubsection{Summary and Outlook}
The extension of alignment techniques from T2I to diverse domains like video, 3D, and molecule generation marks a significant evolution, demonstrating the versatility of core paradigms like RLHF and DPO. A clear pattern has emerged: DPO and its variants are rapidly adopted across modalities due to their simplicity, while RLHF offers powerful control but is hampered by the immense challenge of creating accurate, domain-specific reward models. The central trade-off between implementation simplicity and nuanced control is thus amplified by cross-domain challenges, primarily severe preference data scarcity and the complexity of modeling domain-specific objectives (e.g., temporal consistency or binding affinity). The future of non-T2I alignment hinges on overcoming this data bottleneck, likely through standardized, cross-domain benchmarks and more universal reward frameworks. This will pave the way for robust, and perhaps even hybrid, alignment techniques tailored to the unique properties of video, geometric, and other complex data types.}

\subsection{Open Challenges and Research Frontiers in Diffusion Alignment}
\label{sec:align_diffusion_challenges}
In this subsection, we discuss open challenges and research frontiers in diffusion alignment, focusing on the nuanced comparison between dominant paradigms, insights from LLM alignment, and challenges unique to the diffusion models.

\subsubsection{The RLHF vs. DPO Debate: A Nuanced Comparison}
The choice between the two dominant training-based alignment paradigms, RLHF and DPO, is not a simple matter of superiority but rather a complex trade-off involving training stability, sample efficiency, and robustness to distribution shifts. 

RLHF, which uses a reward model as proxy for human preferences, allows for complex and nuanced reward functions. However, it often suffers from high variance and inefficient sample usage during the reinforcement learning phase. In contrast, DPO simplifies the training pipeline by eliminating the need for an explicit reward model, optimizing the policy directly on preference data. This can lead to more stable training but introduces its own vulnerabilities. Specifically, recent comprehensive studies suggest that a well-tuned PPO can outperform DPO, especially for complex tasks where DPO's performance may degrade due to distribution shifts between the preference data and the policy model~\citep{pmlr-v235-xu24h}. This highlights a critical weakness of DPO: its performance is highly sensitive to the alignment between the training data distribution and the model's evolving policy.

Theoretical analysis further deepens this trade-off, indicating that the optimal choice depends on the relative learning difficulty of the reward function versus the optimal policy. RLHF may be more sample-efficient when the reward model is easier to learn than the policy, whereas DPO holds an asymptotic advantage when the true reward function is exceptionally complex~\citep{pmlr-v235-nika24a}.

Beyond these training-based methods, test-time approaches align models during inference by adjusting inputs, noise, or internal mechanisms, thus avoiding costly fine-tuning. These methods are efficient and easy to deploy but may lack the precision needed for complex alignment tasks~\citep{li2025test}. The entire landscape is rapidly evolving from static, offline training towards more dynamic online and iterative preference learning, which promises to better adapt to feedback and overcome the limitations of fixed preference datasets~\citep{pmlr-v235-xiong24a,pmlr-v235-calandriello24a}.

\subsubsection{Cross-Domain Insights: Adapting Innovations from LLM Alignment}
The human alignment of diffusion models, while nascent, has benefited from adapting techniques pioneered for LLMs. For instance, \citet{wallace2024diffusion} successfully extended DPO~\citep{rafailov2023direct} to create Diffusion-DPO, and \citet{li2024aligning} adapted KTO~\citep{ethayarajh2024kto} to produce Diffusion-KTO. These successes suggest that other advanced LLM alignment methods, such as IPO, ORPO, and PRO (as discussed in \cref{sec:align_without_reward}), are promising candidates for adaptation.

Indeed, the frontier of LLM alignment is rapidly expanding. For instance, f-DPO~\citep{wang2023beyond} generalizes DPO to a broader family of f-divergences for better diversity control, and GPO~\citep{tang2024generalized} provides a unified framework encompassing DPO and IPO. Other novel approaches like BOND~\citep{sessa2024bond}, which distills a policy from a best-of-N sampling distribution, and techniques that derive dense, token-level rewards for free~\citep{chan2024dense}, also represent promising avenues. While these methods have advanced LLM alignment, their transfer to diffusion models is not guaranteed.

A key challenge lies in the fundamental architectural differences. Many LLM alignment techniques leverage the model's auto-regressive, next-token prediction structure in a discrete token space. For example, SimPO~\citep{meng2024simpo} achieves strong performance in LLMs by using a sequence's average log probability as a reference-free implicit reward. Creatively mapping such sequence-based concepts to the iterative, continuous, and high-dimensional denoising process of diffusion models remains a core research question.

\subsubsection{Unique Challenges in Diffusion Model Alignment}
Beyond the general challenges of alignment discussed in \cref{sec:relevant_problms}, diffusion models present a unique set of problems stemming from their generative process and multimodal nature. At a high level, the alignment problem in deep learning is fraught with fundamental risks, such as models learning to ``hack'' proxy reward signals or developing misaligned internal goals that deviate from the intended human preferences~\citep{ngo2024the}. In diffusion models, these challenges are amplified and new ones emerge.

First, feedback for T2I models is inherently subjective and multidimensional, spanning image quality, realism, artistic style, and cultural context. This complexity makes reliable AI-generated feedback non-trivial to obtain. Second, the diversity of human preferences intensifies the issue of distributional shift. Most public preference datasets are built using Stable Diffusion variants (see \cref{tab:feedback data}), and applying these datasets to align other models (e.g., Midjourney) can lead to significant misalignment. This is a critical vulnerability for methods like DPO, whose learning dynamics can be skewed by the specific biases of the training data~\citep{im2024understanding}. Third, the iterative nature of the diffusion process means that feedback may need to be incorporated at multiple steps, demanding highly efficient alignment algorithms. Fourth, integrating feedback from various modalities (e.g., visual, textual, numerical) in a coherent manner adds another layer of complexity. Finally, feedback on images is often sparser and noisier than on text, making it difficult for RL algorithms to learn effectively from inconsistent signals.

Beyond these issues, several profound challenges cast a shadow on current alignment strategies:
\begin{itemize}
    \item[\textbf{(1)}] \textbf{Reward Model Overoptimization:} A core issue in RLHF is ``reward hacking'', where optimizing against a fixed, imperfect reward model leads the policy to find exploitative solutions that maximize the proxy score but fail to capture true human preference. This is a predictable phenomenon governed by scaling laws, where performance on the true objective predictably rises and then falls as the policy diverges from its initial state~\citep{gao2023scaling}. In the visual domain, this can manifest as images that are semantically correct but visually distorted or absurd.

    \item[\textbf{(2)}] \textbf{Scalability and Brittleness:} The long-term viability and robustness of current alignment methods are under question. Large-scale studies suggest RLHF may scale less efficiently than pre-training~\citep{hou2024does}. Furthermore, the resulting alignment can be brittle. For instance, safety alignment in T2I models has been shown to ``backfire'', where fine-tuning on benign data can cause suppressed, unsafe concepts to re-emerge, suggesting the initial alignment was not robustly learned~\citep{kim2024safety}. Even low-rank modifications or model pruning can compromise safety alignment, revealing its fragility~\citep{pmlr-v235-wei24f}.

    \item[\textbf{(3)}] \textbf{Security Vulnerabilities:} The reliance on a reward model introduces a new attack surface. Recent work has demonstrated a ``clean-label'' poisoning attack, termed BadReward, where an attacker subtly poisons the preference dataset with seemingly harmless examples. This manipulation corrupts the learned reward model, which in turn steers the diffusion model to generate harmful or undesired content during RLHF fine-tuning~\citep{duan2025badreward}.
\end{itemize}

\section{Benchmarks and Evaluation for Human Alignment of T2I Diffusion Models}
\label{sec:eval} In this section, we first review benchmark datasets and evaluation metrics for human alignment of T2I diffusion models in \cref{sec:benchmark} and \cref{sec:evaluation_metrics}, respectively. We then discuss the associated challenges in \cref{sec:benchmarks_metrics_challenges}.

\subsection{Benchmarks for Human Alignment of T2I Diffusion Models}
\label{sec:benchmark}
The foundation of any alignment technique is the data used to define human preferences. In this subsection, we discuss benchmark datasets for human alignment of T2I diffusion models, categorizing them into three types that reflect the field's maturation: scalar human preference datasets, multi-dimensional human feedback datasets, and AI feedback datasets. This progression highlights a move towards capturing more nuanced and scalable representations of human intent. \cref{tab:feedback data} compares the reviewed benchmark datasets across three aspects: prompts, images, and annotations.

\subsubsection{Scalar Human Preference Datasets}
Early preference datasets primarily provide an overall comparison among images using a single scalar score or a pairwise choice to indicate human preference. \citet{wu2023hps} introduced the HPD v1 dataset, with prompts and images collected from the public Stable Foundation Discord channel. While valuable, this approach inherently captures the preferences of a niche group of experienced Stable Diffusion users, which may not generalize to the broader population. \citet{wu2023hpsv2} later introduced a larger dataset, HPD v2, where the prompts are sourced from COCO Captions~\citep{chen2015microsoft} and the ChatGPT-cleansed DiffusionDB~\citep{wang2022diffusiondb}. Notably, HPD v2 includes images generated by nine different generative models, including diffusion models, GANs, and auto-regressive-based models, resulting in a higher degree of diversity. The pairwise image preferences in HPD v2 are derived from the preference rankings of 57 employed annotators over the generated images.

To capture more authentic, in-the-wild preferences, \citet{kirstain2023pick} developed a web application to build the Pick-a-Pic v1 dataset, collecting prompts and preferences over images generated by multiple Stable Diffusion variants from thousands of real users. However, such a collection method may be subject to self-selection bias, as the user base of a specific application may not be fully representative. \citet{xu2023imagereward} created the ImageRewardDB dataset by using six popular T2I generative models to generate images based on a diverse selection of prompts from DiffusionDB~\citep{wang2022diffusiondb}. They implemented a three-stage annotation pipeline in which hired annotators annotate prompts, rate text-image pairs, and rank images. This pipeline provides more detailed human preference feedback, capturing aspects such as fidelity, image-text alignment, and overall quality.

Beyond explicit human preferences from annotators regarding image fidelity and image-text alignment, \citet{isajanyan2024social} introduced the Picsart Image-Social dataset, which captures social popularity for creative editing purposes as an implicit and novel dimension of human preferences. Instead of relying on explicit annotations, they utilized editing behaviors (e.g., how often an image is ``remixed'' by others) from the online visual creation and editing platform Picsart to curate this dataset. While this provides a unique, large-scale signal, it is important to note that such a proxy may reflect creative utility or "remixability" more than pure image quality or prompt alignment.

\noindent
\subsubsection{Multi-dimensional Human Feedback Datasets}
Recognizing the limitations of a single preference score, multi-dimensional human preferences~\citep{zhang2024learning} and rich human feedback~\citep{liang2024rich} have been shown to be effective in improving T2I generations. Specifically, motivated by the observation that human preference results vary when evaluating images across different aspects, \citet{zhang2024learning} constructed the MHP dataset. This dataset was created using a balanced and refined prompt set from four sources and nine different T2I diffusion models to generate images with various resolutions and aspect ratios. In particular, 210 crowd-sourced annotators were employed to provide preference choices over image pairs across four dimensions: aesthetics, detail quality, semantic alignment, and overall assessment. Similarly, \citet{liang2024rich} sampled a diverse and balanced subset of image-text pairs from the Pick-a-Pic dataset. They then constructed the RichHF-18K dataset, which provides enriched feedback signals. Specifically, they marked implausible or misaligned image regions, annotated which words in the text prompt were missing or misrepresented in the corresponding image, and provided four fine-grained scores, including plausibility, image-text alignment, aesthetics, and overall quality, on a 5-level Likert scale.

\noindent
\subsubsection{AI Feedback Datasets}
Scaling up human feedback datasets is prohibitively expensive due to the high cost of human annotation. This has motivated researchers to explore AI feedback for constructing preference datasets. \citet{peng2025dreambench} introduced DreamBench++, a benchmark for personalized image generation that uses GPT-4o for automated evaluation aligned with human preferences, focusing on concept preservation and prompt following. Similarly, \citet{wu2024multimodal} created the VisionPrefer dataset using multimodal large language models, specifically GPT-4 Vision. The annotations include scalar scores, preference rankings, and rationales for the annotations across four aspects: prompt-following, fidelity, aesthetics, and harmlessness. They then trained a reward model, VP-Score, based on VisionPrefer. VP-Score demonstrates comparable performance to reward models trained on human preference datasets in predicting human preferences and aligning T2I diffusion models with these preferences.

\begin{table}[t]
\caption{Comparison of existing feedback datasets for T2I diffusion models.}
\resizebox{0.99\linewidth}{!}{
\centering
\begin{tabular}{c|cc|cc|ccc} 
\toprule
Feedback Dataset $\rightarrow$ Reward Model & Prompt Source& Prompt Count& Image Generation Source& Image Count & Annotator Info. & Annotation Count \\ 
\midrule

\makecell{HPD v1 $\rightarrow$ HPS v1 \\~\citep{wu2023hps}}&  \makecell{Stable Foundation\\ Discord channel} &  25,205&  Stable Diffusion &  98,807&  \makecell{2659 \\ experienced users} &25,205 \\ 

\midrule
\makecell{HPD v2 $\rightarrow$ HPS v2\\~\citep{wu2023hpsv2}}&  \makecell{COCO Captions +  \\DiffusionDB}&  107,915&  \makecell{9 models + \\ COCO images}&  433,760&  \makecell{57 \\ employed annotators} & 798,090\\ 

\midrule
\makecell{Pick-a-Pic v1 $\rightarrow$ PickScore \\~\citep{kirstain2023pick}} &  Real users&  37,523&  \makecell{Stable Diffusion \\ variants} &  1,169,494&  \makecell{6,394 \\ web app users} &584,747\\ 

\midrule
\makecell{ImageRewardDB $\rightarrow$ ImageReward\\~\citep{xu2023imagereward}} &  DiffusionDB &  8,878&  6 models&  273,784&  \makecell{Annotation \\ company} &136,892\\ 

\midrule
\makecell{MHP $\rightarrow$ MPS  \\~\citep{zhang2024learning}}&  \makecell{PromptHero + \\ DiffusionDB + \\ KOLORS + GPT4}&  66,389 &  9 models&  607,541& \makecell{210 crowd-sourced \\ annotators} &918,315\\ 

\midrule
\makecell{RichHF-18K $\rightarrow$ RAHF \\~\citep{liang2024rich}}&  Pick-a-Pic v1&  17,760&  Pick-a-Pic v1&  35,520& \makecell{27 \\ trained annotators} &17,760\\ 

\midrule
\makecell{Picsart Image-Social $\rightarrow$ Social Reward \\~\citep{isajanyan2024social}}&  \makecell{Social platform \\ user prompts}&  104 K &  \makecell{Several \\ in-house models} &  1.7 M & \makecell{1.5 M \\ users} & 3M \\ 

\midrule
\makecell{VisionPrefer $\rightarrow$ VP-Score \\~\citep{wu2024multimodal}}&  \makecell{DiffusionDB}&  179 K &  \makecell{Stable Diffusion \\ variants} &  0.76 M & \makecell{GPT-4 Vision} & 1.2 M \\ 

\bottomrule 
\end{tabular}
} 

\label{tab:feedback data}
\end{table}

\subsection{Evaluation for Human Alignment of T2I Diffusion Models}
\label{sec:evaluation_metrics} 
The benchmarks described previously enable the development of diverse evaluation paradigms. In this subsection, we review the primary methods for assessing alignment, starting with the evaluation of the reward models themselves in \cref{sec:eval_reward_model}, and then moving to the evaluation of the final T2I diffusion models in \cref{sec:eval_t2i_model}.

\subsubsection{Evaluation for Reward Models}
\label{sec:eval_reward_model} 
To evaluate the performance of reward models in predicting human preference, the classical metric used is pairwise preference prediction accuracy. To calculate this accuracy, the reward model is first used to score a pair of images with the same prompt. The accuracy is then determined by the ratio of cases where the reward model assigns a higher score to the image-text pair preferred by humans on the test set. While high accuracy on a given benchmark is a necessary indicator, it is not sufficient, as a model can overfit to the benchmark's specific data distribution and fail to generalize to novel, out-of-distribution prompts. 

\cref{tab:feedback data} delineates the mapping between prominent feedback datasets and their corresponding reward models. Building on this, \cref{table:reward_models} presents a comparative analysis of human preference prediction accuracy for nine models across five benchmarks. The results reveal a consistent pattern of benchmark specialization: models such as MPS, PickScore, and Social Reward achieve state-of-the-art performance predominantly on their native datasets. {This finding strongly indicates that a model's predictive power is highly contingent upon the specific data distribution of its training benchmark. Furthermore, the overall prediction accuracies are modest, with most remaining below $80\%$, suggesting that robustly capturing general human preferences remains a significant open challenge.}

In addition to predicting overall human preference on generated images from the same prompt~\citep{wu2023hps, wu2023hpsv2, kirstain2023pick, xu2023imagereward, wu2024multimodal, isajanyan2024social}, novel reward models have been proposed to predict multi-dimensional preferences~\citep{zhang2024learning}, detect implausible or misaligned regions, and identify misaligned keywords~\citep{liang2024rich}. As a result, distinct metrics have been developed for evaluation, including the correlation between Elo ratings~\citep{elo1978rating} of real users and reward models~\citep{kirstain2023pick}, the correlation between the win ratio of reward models and humans~\citep{kirstain2023pick, zhang2024learning}, and metrics like NSS, KLD, AUC-Judd, SIM, and CC~\citep{bylinskii2018different} for evaluating saliency heatmaps~\citep{liang2024rich}.

\begin{table}[t]
\caption{Comparison of different reward models for human preference evaluation. The pairwise preference prediction accuracy ($\%$) is reported on ImageRewardDB, HPD v2, MHP, Pick-a-Pic v1, and Picsart Image-Social dataset. The \textbf{bold} results indicate the best result on each dataset. The results with no mark, $*$, and $**$ are from \citet{zhang2024learning}, \citet{wu2024multimodal}, and \citet{isajanyan2024social}, respectively.}
\centering
\begin{tabular}{c|ccccc}
\toprule
  & ImageRewardDB & HPD v2 & MHP & Pick-a-Pic v1 & Picsart Image-Social \\ 

\midrule
\makecell{CLIP score~\citep{radford2021learning}} & 54.3 & 71.2 & 63.7 & 60.8* & 51.9 \\ 

\midrule
\makecell{Aesthetic score~\citep{schuhmann2022laion}} & 57.4 & 72.6 & 62.9 & 56.8* & 55.3 \\ 

\midrule
\makecell{HPS v1~\citep{wu2023hps}} & 61.2 & 73.1 & 65.5 & 66.7* & - \\ 

\midrule
\makecell{HPS v2~\citep{wu2023hpsv2}} & 65.7 & 83.3 & 65.5 & 67.4* & 59.4 \\ 

\midrule
\makecell{PickScore~\citep{kirstain2023pick}} & 62.9 & 79.8 & 69.5 & \textbf{70.5*} & 62.6 \\ 

\midrule
\makecell{ImageReward~\citep{xu2023imagereward}} & 65.1 & 70.6 & 67.5 & 61.1* & 60.5 \\ 

\midrule
\makecell{MPS~\citep{zhang2024learning}} & \textbf{67.5} & \textbf{83.5} & \textbf{74.2} & - & - \\ 

\midrule
\makecell{VP-Score~\citep{wu2024multimodal}} & 66.3* & 79.4* & - & 67.1* & - \\ 

\midrule
\makecell{Social Reward~\citep{isajanyan2024social}} & - & - & - & - & \textbf{69.7} \\ 

\bottomrule
\end{tabular}
\label{table:reward_models}
\end{table}

\subsubsection{Evaluation for T2I Diffusion Model}
\label{sec:eval_t2i_model}\
\\
\noindent
\textbf{Model Evaluation Prompts.} To evaluate T2I diffusion models, it is essential to collect a representative set of prompts for image generation that aligns with the evaluation goals. Various prompt datasets are available for T2I model evaluation in the context of human alignment. For example, \citet{kirstain2023pick} used prompts from MS-COCO~\citep{lin2014microsoft} and Pick-a-Pic v1 for evaluation, while \citet{xu2023imagereward} selected prompts from DiffusionDB~\citep{wang2022diffusiondb} and MT Bench~\citep{petsiuk2022human}. \cref{tab:feedback data} outlines the prompt sources for each feedback dataset, highlighting different motives for image generation, such as real user intention~\citep{kirstain2023pick}, challenging multi-task prompts~\citep{petsiuk2022human}, and social popularity~\citep{isajanyan2024social}. Consequently, we recommend that the community employ suitable prompts when assessing the performance of T2I diffusion models across different evaluation aspects.

\noindent
\textbf{Image Quality.} The Inception Score (IS)~\citep{salimans2016improved} and Fréchet Inception Distance (FID)~\citep{heusel2017gans} are the most widely adopted metrics for measuring image quality without considering the text prompt. These metrics utilize features extracted from a pre-trained image classifier, typically the Inception-V3 model~\citep{szegedy2016rethinking}, to evaluate the fidelity and diversity of generated images. However, their primary limitation is that they do not account for the text prompt and their correlation with human perceptual judgment is imperfect, rendering them insufficient for a comprehensive assessment of alignment.

\noindent
\textbf{Human Preference Evaluation.} Reward models can serve as metrics for human preference, allowing comparisons between various T2I generative models based on their reward scores, or for monitoring the training process of aligning models with human preferences. Typically, reward scores will show an increasing trend when models are fine-tuned using RLHF methods (see \cref{sec:align_rlhf_diffusion}) or through DPO approaches (see \cref{sec:align_dpo_diffusion}) with feedback datasets. This increasing trend indicates improved alignment with human preferences, as measured by the reward models. A critical caveat here is the risk of ``evaluation hacking'' or circular reasoning: using a reward model to evaluate a policy that was optimized using that same model (or a very similar one) can lead to inflated scores that do not reflect true gains in alignment. Notably, most reward scores account for the text prompt, often computed as a scaled cosine similarity, with the exception of metrics like the aesthetic score~\citep{schuhmann2022laion}, which measures aesthetics independently. For a recent review of T2I evaluation, see~\citet{hartwig2024evaluating}.

\noindent
{
\textbf{Fine-grained Evaluation.} The automated evaluation metrics introduced above offer a holistic measure of quality but often lack interpretability. To address this, recent works focus on fine-grained, instance-level analysis to better reflect the diverse capabilities of T2I models. These efforts can be categorized into several key areas:
\begin{itemize}
    \item \textbf{Compositional Reasoning:} Benchmarks like GENEval~\citep{ghosh2023geneval} and GenAI-Bench~\citep{lin2024evaluating} assess a model's ability to handle complex prompts involving multiple objects, attributes, and spatial relationships. VQAScore~\citep{lin2024evaluating} probes this by using a visual question-answering model to verify compositional correctness with a binary question.
    \item \textbf{Visual Reasoning and Social Bias:} DALL-Eval~\citep{cho2023dall} was designed to probe models' commonsense reasoning skills (e.g., object counting, spatial relations) and to quantify social biases related to protected attributes like gender and skin tone.
    \item \textbf{Multi-faceted Skill Assessment:} Comprehensive benchmarks like HEIM~\citep{lee2023holistic} and VPEval~\citep{goyal2023visual} evaluate models across a wide spectrum of skills (e.g., text understanding, photorealism, counting) and provide more interpretable, explanatory results.
    \item \textbf{LLM-based Evaluation:} LLMScore~\citep{lu2023llmscore} leverages the descriptive power of LLMs to first generate a textual caption for an image and then score its alignment with the original prompt, providing a human-like rationale for its assessment.
    \item \textbf{Style Attribution:} Other work has focused on evaluating more abstract concepts like artistic style, for instance, by measuring a model's ability to attribute and match the style of specific artists~\citep{somepalli2024measuring}.
\end{itemize}
}

\subsection{Challenges in Benchmarking and Evaluation for T2I Diffusion Models}
\label{sec:benchmarks_metrics_challenges}
While the infrastructure for alignment is maturing, it faces profound challenges that limit the reliability and scope of current research. These challenges fall into two interconnected categories: those related to benchmark construction and those inherent to evaluation methodologies.

\noindent
\textbf{Challenges in Benchmark Construction.} Creating a benchmark that truly represents human preference is fraught with difficulties. First, human preferences are inherently subjective, diverse, and dynamic. Most current datasets, while striving for diversity in prompts and models, are often annotated by a limited number of individuals, failing to capture the full spectrum of human taste. This can lead to models aligned with a homogenous, averaged preference rather than a pluralistic one, as current alignment frameworks often enforce uniformity through strict rubrics, thereby suppressing the natural diversity of human opinions~\citep{chen2025pal}. Consequently, the claimed diversity of datasets must be rigorously measured, not just asserted~\citep{zhao2024position}. Second, all existing benchmarks are static snapshots. They cannot account for the evolving nature of human preferences over time, a problem that might ultimately require continual learning approaches to solve~\citep{10444954}.

\noindent
\textbf{Challenges in Evaluation Methodologies.} The metrics used for evaluation face their own set of challenges. The lack of standardization in evaluation prompts complicates consistent comparison across studies; for instance, auto-generated prompts may lack diversity and objectivity compared to those sourced from professionals~\citep{lin2024evaluating}. Establishing widely adopted, standardized evaluation protocols remains an open problem. Furthermore, the increasing reliance on multimodal large language models (MLLMs) as automated evaluators is a double-edged sword. While scalable, these MLLMs can suffer from their own issues, such as positional bias, and their evaluation quality is capped by the capabilities of the underlying model, potentially propagating these flaws into the metrics themselves~\citep{peng2025dreambench, wu2024multimodal}.

Finally, a superordinate challenge is that current metrics struggle to assess abstract yet crucial qualities like creativity~\citep{10.1145/3664595}. This points to the significant risk of an ``alignment tax'': the phenomenon where excessive optimization for specific, measurable alignment goals (e.g., prompt following, safety) inadvertently suppresses a model's creativity, diversity, and overall utility. Understanding and mitigating this trade-off is critical for developing genuinely helpful and innovative generative models.

\section{Future Directions for Human Alignment of T2I Diffusion Models}
\label{sec:future}
In this section, we outline three promising future directions for human alignment of T2I diffusion models that can inspire further advancements in the area.

\noindent
\textbf{Preference Learning with Inconsistent and Multidimensional Human Feedback.}
A critical limitation of current alignment paradigms is their tendencies to collapse diverse human preferences into a single, monolithic reward function. This approach fails to capture the rich, often conflicting, and subjective nature of human values. {This challenge is particularly acute for diffusion models, as visual content—encompassing aesthetics, artistic style, and cultural context—is inherently more subjective and ambiguous than text. A major future direction is therefore to develop systems that embrace \textit{pluralistic alignment}~\citep{sorensen2024roadmap}.} Instead of seeking a single best output, the goal is to model a distribution of desirable outputs that reflect a wide range of viewpoints.

This requires moving beyond simple reward modeling. One approach is to draw upon the formalisms of \textit{social choice theory} to aggregate diverse feedback in a principled manner, leading to paradigms like Reinforcement Learning from Collective Human Feedback (RLCHF)~\citep{10.5555/3692070.3692441}. Another is to fundamentally change the optimization objective. For instance, Nash Learning from Human Feedback (NLHF) learns a preference relation instead of a scalar reward, which is more robust to the non-transitive or cyclical preferences that naturally arise in group feedback~\citep{munos2024nash}. To explicitly handle distinct preference groups, MaxMin-RLHF identifies different user sub-populations and optimizes for the worst-case reward among them, ensuring the model does not cater only to the majority~\citep{chakraborty2024maxminrlhf}. {Applying these advanced aggregation mechanisms to diffusion models presents unique opportunities and challenges, requiring novel algorithms that can navigate optimization in a high-dimensional, continuous generation space while handling the sparse nature of image-based feedback. For more dynamic control, methods like Rewards-in-Context~\citep{yang2024rewards}, which allow users to specify multi-objective preferences at inference time, represent a promising path toward achieving not just learned, but \textit{controllable} pluralistic alignment, striking a balance between general preference satisfaction and personalized generation.}

\noindent
\textbf{Data-centric Preference Learning.}
Current preference learning approaches for diffusion models typically rely on supervised learning with large-scale human-annotated datasets. While diffusion models can generate content at low cost, obtaining human feedback on these outputs remains expensive and slow, particularly in visual domains. This data bottleneck is exacerbated by findings that current alignment techniques may not scale efficiently, requiring ever-larger datasets for diminishing gains~\citep{hou2024does}. To address this, researchers can explore preference learning with minimal or even zero human-annotated data.

Two potential approaches can mitigate this dependency. First, while AI feedback is a scalable alternative, it carries risks of inheriting biases, lacking diversity, and potential model collapse from training on synthetic data~\citep{shen2024finetuning}. This area requires further exploration to enhance reliability. Second, a promising direction is using AI-generated paired samples with known preference relations. For instance, if we know that Algorithm A produces better results than Algorithm B based on a specific alignment metric (e.g., photorealism or prompt-following), we can use these paired samples to curate a large-scale, low-cost preference dataset to improve that metric. For better preference diversity and reliability, we may choose multiple algorithms to generate paired samples and use the alignment metric score to remove those pairs without significant preference score differences. Such data-centric strategies significantly reduce reliance on direct human annotation. However, their efficacy is still bound by predefined external metrics. A more ambitious goal is to enable models to develop an intrinsic understanding of preference, which leads to our final research frontier: self-alignment.

\noindent
\textbf{Self-Alignment of Diffusion Models.}
Currently, methods for aligning diffusion models rely on intensive external supervision. We propose a forward-looking direction, \textit{self-alignment}, built on the hypothesis that large diffusion models, trained on vast, high-quality data, already possess implicit prior knowledge of human preferences (e.g., aesthetics, realism) but lack a mechanism to express it explicitly ~\citep{burns2023weak,sun2023principle,pang2024self}. The goal is to unlock this latent capability, enabling a model to align itself with minimal or no external feedback. This would allow a model to act as its own reward function for RLHF or as an AI annotator for DPO, creating a self-improving loop that could eventually surpass human capabilities. This vision has been discussed for LLMs but is underexplored for diffusion models.

{Achieving self-alignment may follow two complementary paths:}
{\begin{itemize}
    \item \textbf{Unlocking Inherent Judgment:} The first path focuses on extracting and amplifying the model's existing, but latent, evaluative abilities. Initial work on \textit{diffusion classifiers}~\citep{li2023your,clark2024text,chen2024robust,qi2024simple} has shown that these models can judge text-image alignment. Future work could go further by probing internal representations, such as intermediate attention maps~\citep{chefer2023attend} or influence functions~\citep{koh2017understanding,yang2023dataset}, to construct a more comprehensive, self-generated reward signal.
    \item \textbf{Fostering Self-Reflective Generation:} The second path aims to embed alignment into the generation process itself. Recent explorations into equipping models with a Chain-of-Thought (CoT) \citep{wei2022chain}-like reasoning process are a step in this direction~\citep{guo2025can}. By generating content step-by-step and using internal signals to verify and refine each stage, the model can engage in a form of self-correction, promoting more coherent and aligned outputs without external guidance.
\end{itemize}}
{The realization of self-alignment would mark a paradigm shift from models being passively aligned to becoming active agents in their own refinement. This could not only resolve the data bottleneck but also unleash new levels of creative and intelligent generation.}

\section{Summary}
\label{sec:conclusion}
In this paper, we have presented a comprehensive review of the alignment of diffusion models. We explored recent advances in diffusion models, elucidated fundamental concepts of human alignment, and discussed various techniques for enhancing the alignment of diffusion models, as well as extending these techniques to tasks beyond T2I generation. Additionally, we outlined the benchmark datasets and evaluation metrics critical for assessing T2I diffusion models. Looking ahead, we identified a series of profound challenges and several promising directions for future research. We hope that this work not only highlights recent advancements and existing gaps in diffusion alignment but also inspires and guides future alignment research of diffusion models.
\begin{acks}
This work was supported by the National Natural Science Foundation of China under Grant No. 62506317.
\end{acks}

\bibliographystyle{ACM-Reference-Format}
\bibliography{alignment}


\end{document}